\documentclass[a4paper,fleqn]{cas-dc}

\usepackage[numbers]{natbib}

\usepackage{framed,multirow}

\usepackage{amssymb}
\usepackage{latexsym}

\usepackage{url}
\usepackage{xcolor}

\usepackage{subcaption}

\usepackage{hyperref}
\usepackage{graphicx}
\usepackage{amsmath}
\usepackage{amssymb}
\usepackage{tabularx,longtable,multirow}
\usepackage{color}
\usepackage{xcolor}
\usepackage{tabularx}
\usepackage{natbib}
\usepackage{supertabular}
\usepackage{epsfig}
\usepackage{graphicx}
\usepackage{amssymb}
\usepackage{colortbl}
\usepackage{relsize}
\usepackage{multirow}
\usepackage{float}
\usepackage{mathtools}

\usepackage[nolist]{acronym}
\usepackage[noend]{algpseudocode}
\usepackage[ruled,commentsnumbered]{algorithm2e}
\usepackage{booktabs}

\usepackage{makecell}
\usepackage{hhline}
\usepackage{array, spreadtab}
\usepackage{collcell,booktabs,siunitx}
\usepackage{xparse}
\usepackage{fp}
\usepackage{pgfplotstable}

\def\tsc#1{\csdef{#1}{\textsc{\lowercase{#1}}\xspace}}
\tsc{WGM}
\tsc{QE}
\tsc{EP}
\tsc{PMS}
\tsc{BEC}
\tsc{DE}
\usepackage{placeins}
\ExplSyntaxOn
\cs_gset:Npn \__first_footerline: {}
\ExplSyntaxOff
\begin{document}
\let\WriteBookmarks\relax
\shorttitle{KANMultiSign for Human Pose Animation}
\shortauthors{G. Du et~al.}

\title [mode = title]{KAN Text to Vision? The Exploration of Kolmogorov-Arnold Networks for Multi-Scale Sequence-Based Pose Animation from Sign Language Notation}                      



\author[1]{Guanyi Du}[style=chinese]
\author[1]{Lintao Wang}[style=chinese]
\author[2]{Kun Hu}[style=chinese]
\author[3]{Ziyang Wang\corref{cor1}}[style=chinese]

\affiliation[1]{organization={School of Computer Science, The University of Sydney},
                addressline={Camperdown Campus, NSW 2006},
                city={Sydney},
                postcode={2006},
                country={Australia}}

\affiliation[2]{organization={School of Science, Edith Cowan University},
                addressline={270 Joondalup Drive},
                city={Joondalup},
                postcode={6027},
                state={Western Australia},
                country={Australia}}
\affiliation[3]{organization={School of Computer Science and Digital Technologies, Aston University},
                addressline={Aston Triangle},
                city={Birmingham},
                postcode={B4 7ET},
                country={United Kingdom}}

\cortext[cor1]{Corresponding author}

\begin{abstract}
Sign language production from symbolic notation offers a scalable route to accessible sign animation. We present KANMultiSign, a multi-scale sequence generator that translates HamNoSys notation into two-dimensional human pose sequences. Our framework makes two complementary contributions. First, we introduce a coarse-to-fine generation strategy with multi-scale supervision: the model is first guided by an intermediate body--hand--face scaffold to encourage global structural coherence, and then refines fine-grained hand articulation to improve finger-level detail. Second, we investigate integrating Kolmogorov--Arnold Network modules into a Transformer backbone, using learnable univariate function primitives to model the highly non-linear mapping from discrete phonological symbols to continuous body kinematics with a compact parameterization. Experiments on multiple public corpora spanning Polish, German, Greek, and French sign languages show consistent reductions in dynamic time warping based joint error compared with a strong notation-to-pose baseline, while using substantially fewer parameters. Controlled ablations further indicate that KAN-based variants substantially reduce parameter count while maintaining competitive performance when coupled with multi-scale supervision, rather than serving as the main driver of accuracy gains. These findings position multi-scale supervision as the key mechanism for improving notation-conditioned pose generation, with KAN offering a compact alternative for efficient modeling. Our code will be publicly available.
\end{abstract}


\maketitle

\section{Introduction}
\label{sec:intro}

Sign languages are fully fledged visual-gestural languages used by deaf and hard‑of‑hearing communities worldwide \cite{who2021,sandler2006sign,un2022}. According to the World Health Organisation (WHO), over 430 million people worldwide have disabling hearing loss requiring rehabilitation, with this number continually increasing. Among them, those with profound hearing loss often use sign language for communication, and this total number is projected to increase to over 700 million by 2050~\cite{who2021}. Unlike spoken languages, meaning in sign language emerges from the coordinated dynamics of manual articulators (hands and arms) together with non-manual cues (facial expressions, gaze, and upper‑body pose). Compared to spoken languages, sign language is more accessible for deaf people, because it does not rely on their
familiarity with the written or spoken forms of spoken languages \cite{rastgoo_all_2022}. Bridging communication between signing and non-signing populations motivates technologies for Sign Language Production (SLP), i.e., the automatic generation of sign motion from symbolic or textual inputs.

\begin{figure}
    \centering
    \includegraphics[width=0.8\linewidth]{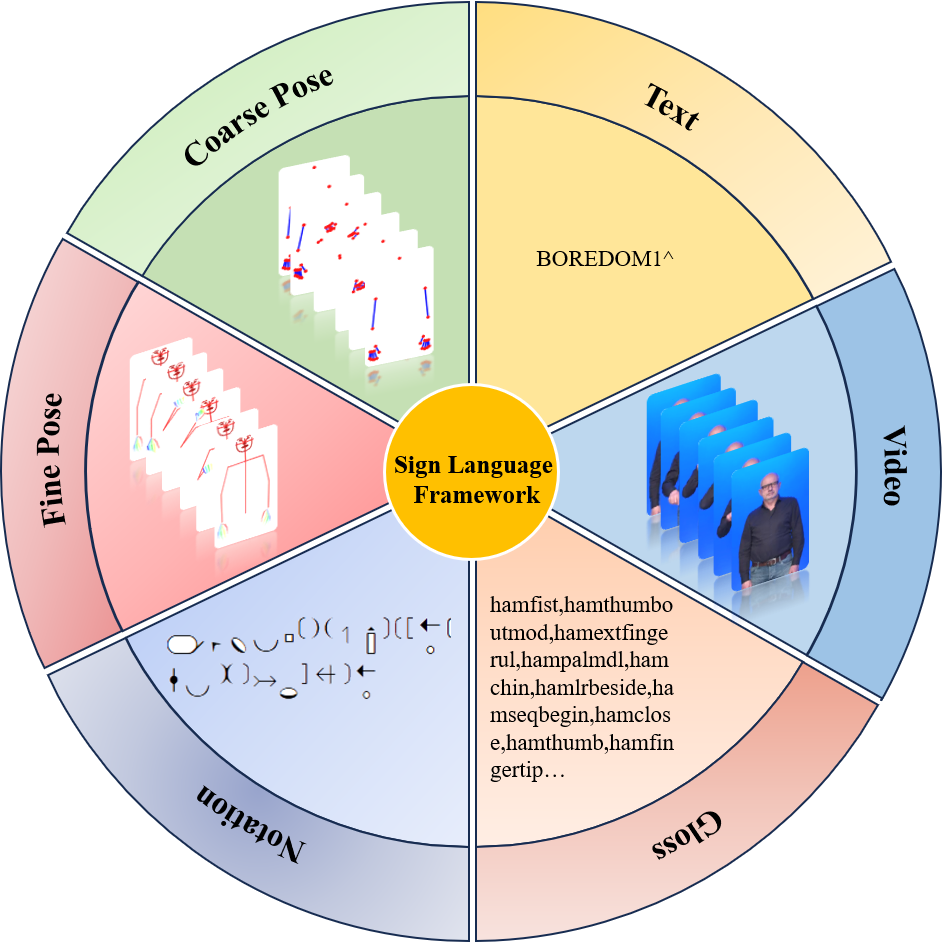}
    \caption{Data representations used in the KANMultiSign study.  The process involves multiple representations of a sign: the raw video, a high-level text label, and symbolic transcriptions such as gloss and HamNoSys notation. Our model translates a symbolic input into motion through a multi-scale approach, first generating a simplified Coarse Pose to capture global structure, which is then refined into the detailed Fine Pose with full articulation.}
    \label{fig:datademo}
\end{figure}

\begin{figure*}[htbp]
\centering
\includegraphics[width=2\columnwidth]{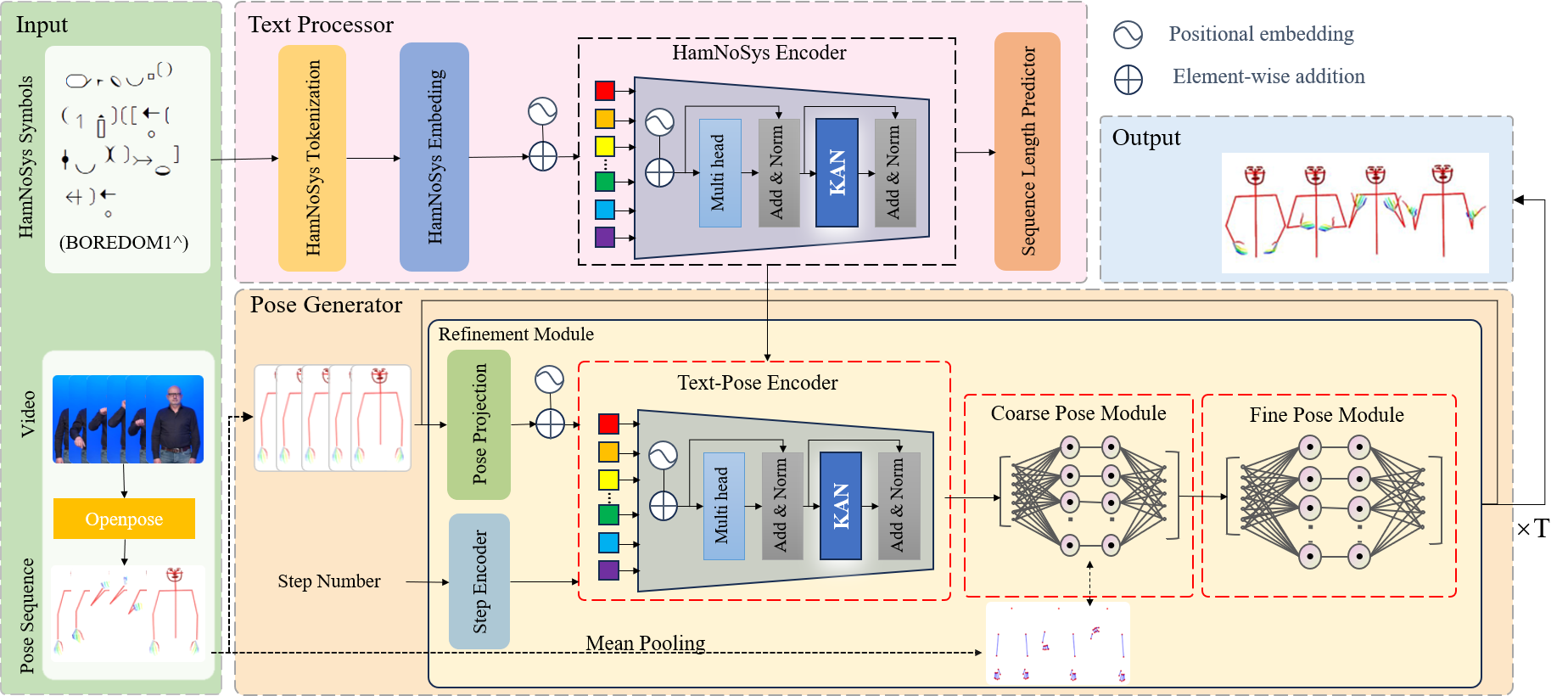}
\caption{The Architecture of the KANMultiSign Framework. First, a Text Processor converts an input HamNoSys sequence into contextual embeddings and predicts the sequence length. Second, a Pose Generator iteratively synthesizes the animation. Within each refinement step, a multi-scale pathway predicts a simplified Coarse Pose to guide the generation of the final Fine Pose.}
\label{fig:1}
\end{figure*}
A practical path to SLP is to leverage linguistic \emph{notation systems} that compactly encode phonological components of signs. Among these, the Hamburg Notation System (HamNoSys) provides language‑agnostic symbols for hand shape, location, orientation, and movement, and has been widely adopted in corpus creation and generation tools \cite{walsh2022changing,zifan2022signwriting}.  These systems provide a structured
way to describe sign languages into visual symbols, offering a universal framework that supports the accurate depiction of signs across different languages. The use of such notation systems is critical not only for educational and documentation purposes but also for technological applications aiming to translate and animate sign languages through computational models. Compared to glosses-language-specific word surrogates that under‑specify non-manuals—and pictographic SignWriting, HamNoSys offers a precise, linear representation that is amenable to machine learning and cross-lingual modeling.

Recent research has begun translating symbolic inputs into motion. Text‑to‑motion models condition on natural language to synthesize human poses or gestures \cite{ahuja2019language2pose,ghosh2021synthesis,petrovich2022temos,tevet2022human,guo2022generating}, while SLP systems map glosses or HamNoSys sequences to signing motion or pose \cite{stoll2020text2sign,saunders2020progressive,arkushin2023ham2pose}. 
Despite progress, a core challenge remains: effectively modeling the profoundly complex and non-linear mapping from an abstract symbolic representation like HamNoSys to the continuous, high-dimensional kinematics of human motion. Standard architectures often rely on parameter-heavy MLP blocks and struggle to capture this mapping efficiently. This results in symptomatic failures such as imprecise manual articulation, particularly in the fine-grained motion of the fingers, and poor generalization when training data are limited.

We address these challenges with \textit{KANMultiSign}, a \emph{multi-scale} HamNoSys-to-pose generator built around a coarse-to-fine training strategy. The central idea is to first guide generation with an intermediate body-hand-face scaffold that captures global structure, and then refine this representation into detailed full-body keypoint sequences. Within this framework, we further explore KAN modules as parameter-efficient replacements for Transformer feed-forward sublayers, leveraging their learnable univariate functions to provide a more compact parameterization for the complex, non-linear mappings from symbolic inputs to body kinematics \cite{liu2024kan}, aiming to reduce model size while preserving competitive generation quality. This paper makes the following contributions:

\begin{enumerate}
    \item \textbf{Multi‑scale coarse‑to‑fine modeling.} We introduce a supervisory path that predicts a 25‑part coarse skeleton to encode global structure before refining to 137 keypoints, improving manual articulation while maintaining body coherence.
    \item \textbf{KAN‑augmented sequence generator for SLP.} We propose a HamNoSys‑conditioned architecture that replaces Transformer MLP sublayers with KAN modules, yielding a parameter‑efficient alternative for symbol‑to‑pose generation.
    \item \textbf{Comprehensive evaluation and controlled attribution.} Across four sign language corpora (PJM, DGS, GSL, and LSF), our multi-scale framework improves DTW-MJE on three datasets and remains competitive on the fourth, while KAN-based variants reduce parameters by up to 55\% (1.7M vs.\ 3.8M). Controlled ablations further verify the respective contributions of multi-scale supervision and KAN modules.
\end{enumerate}


\section{Related Work}
\label{sec:related}

\subsection{Notation Systems and Notation-Driven SLP}
Sign languages can be transcribed with several systems that differ in granularity and suitability for computation. \emph{Gloss} offers language-specific word surrogates but under-specifies non-manual markers; \emph{SignWriting} provides rich pictographic detail, but its spatial layout complicates machine parsing; and \emph{HamNoSys} encodes handshape, orientation, location, and movement in a language-agnostic linear stream amenable to modeling \cite{walsh2022changing,zifan2022signwriting,hanke2004hamnosys}. Due to its formalized, linear structure, HamNoSys has become the predominant choice for computational modeling and is the focus of this work. Early SLP relied on rule-based pipelines and avatar animation, e.g., VisiCast \cite{bangham2000virtual}, eSign \cite{zwitserlood2004synthetic}, Dicta-Sign \cite{efthimiou2010dicta}, and JASigning \cite{ebling2016building}, which required extensive linguistic engineering and did not scale gracefully to fine finger articulation or coarticulation effects.

Neural approaches have since begun mapping symbolic inputs to motion. Work from gloss to signing motion/pose leverages sequence models or Transformers to generate continuous pose from gloss strings \cite{stoll2020text2sign,saunders2020progressive}. Of particular relevance, Ham2Pose directly conditions on HamNoSys and predicts 2D pose sequences from OpenPose keypoints, establishing a strong notation-to-pose baseline that we build upon \cite{arkushin2023ham2pose}. Our approach differs by introducing (i) a multi-scale supervisory pathway that first predicts a coarse body–hand–face representation and then refines to finger-level detail, and (ii) KAN-based feed-forward replacements to improve parameter efficiency and expressivity.

\subsection{Text-/Gloss-to-Motion Outside SLP}
Beyond sign, text-conditioned human motion synthesis explores mapping free-form language descriptions to full-body motion. Representative methods include recurrent or Transformer architectures for language-to-gesture/motion \cite{ahuja2019language2pose,ghosh2021synthesis} and generative latent models for text-conditioned sequences \cite{petrovich2022temos,tevet2022human,guo2022generating}. These works highlight two recurring design choices that are also pertinent to SLP: (i) \emph{content hierarchy}, where global body layout is estimated before refining fine-grained details; and (ii) \emph{temporal scheduling}, where iterative refinement or noise schedules improve temporal coherence. KANMultiSign adopts a coarse-to-fine design specialized for HamNoSys and couples it with a length-aware refinement schedule tailored to sign timing.

\subsection{Multi-Scale and Structural Modeling of Human Motion}
Multi-scale skeletal and motion representations improve stability and detail by capturing long-range structure and local articulation. Prior work uses higher-order graph operators or Multi-Scale designs to encode structure at different resolutions ~\cite{jang2022motion, liu2020disentangling}, as well as heuristic or hierarchical decompositions for controllable gesture and hand detail ~\cite{bhattacharya2021speech2affectivegestures, dang2021msr,ghosh2021synthesis}. In parallel, recent sign language video generation works explicitly incorporate fine-grained motion conditions, including detailed pose cues and 3D hand information, to improve articulation fidelity and temporal stability ~\cite{shi2024poseguidedfinegrainedsignlanguage, wang2025advancedsignlanguagevideo, yu2024signavatarslargescale3dsign}. Moreover, structure-aware objectives have been explored for holistic 3D SLP, e.g., masked modeling strategies that explicitly exploit kinematic hierarchy to improve articulation fidelity and temporal stability \cite{10.1145/3776750}. Similar methodological ideas have also been explored in intelligent industrial inspection, where multi-scale modeling, fine-grained structural analysis, and robust learning are important for reliable recognition of subtle local patterns. Although the application differs from sign language production, these studies help contextualize our coarse-to-fine design ~\cite{ZENG2026133294, WU2026101084}. 
Our formulation formalizes a 25-part coarse skeleton (body segments and hand digits) as an intermediate target that supervises global structure before regressing the full 137-keypoint sequence, which we find improves handshape fidelity without sacrificing whole-body coherence.

\subsection{Generative Backbones for Motion}
Denoising diffusion models provide strong priors for high-fidelity synthesis \cite{sohl2015deep,ho2020denoising,dhariwal2021diffusion,saharia2022photorealistic}, and related retargeting/synthesis systems demonstrate convincing human motion generation \cite{chan2019everybody}. However, the iterative sampling process in diffusion models can be computationally heavy for real-time SLP. We therefore adopt a lightweight iterative refinement schedule within a Transformer-style encoder, retaining temporal smoothing benefits while keeping inference efficient.

\subsection{Kolmogorov–Arnold Networks as MLP Alternatives}

The Kolmogorov--Arnold representation theorem motivates approximating multivariate mappings via superpositions of learnable univariate functions, offering an alternative to fixed linear weights in standard MLPs~\cite{kolmogorov1961representation}. Building on this principle, modern Kolmogorov--Arnold Networks (KANs) have been instantiated as practical neural modules with improved parameter efficiency and post-hoc interpretability in a range of settings~\cite{liu2024kan}. 

Recent studies have extended Kolmogorov–Arnold Networks (KANs) beyond their initial demonstrations, applying them to a range of domains including time-series forecasting and anomaly detection ~\cite{xu2024kolmogorovarnoldnetworkstimeseries, zhou2025kanad, Vaca_Rubio_2024}, computer vision recognition/segmentation ~\cite{azam2024suitabilitykanscomputervision, cang2024kanworkexploringpotential, bodner2025convolutionalkolmogorovarnoldnetworks, li2024ukanmakesstrongbackbone}, graph representation learning~\cite{kiamari2024gkangraphkolmogorovarnoldnetworks, bresson2025kagnnskolmogorovarnoldnetworksmeet, decarlo2025kolmogorovarnoldgraphneuralnetworks}. 

Despite these advances, KANs have not been systematically explored in notation-driven sign language production, where the model must map discrete phonological symbols (e.g., HamNoSys) to continuous, high-dimensional kinematics with strict temporal consistency and fine-grained hand articulation. This gap motivates our study of KAN-based FFNs within a notation-driven pose-generation framework that is primarily strengthened by coarse-to-fine multi-scale supervision. In our formulation, the multi-scale pathway addresses global-to-local structural learning, while KAN is investigated as a compact alternative to standard MLP-based FFNs.

\section{Methodology}
\label{sec:methodology}

Given a sign's HamNoSys notation, represented as the symbol sequence 
\(H = (h_1, \dots, h_N)\), the objective of our model is to generate a corresponding sequence 
of 2D human poses, \(P = (p_1, p_2, \dots, p_T)\).
As illustrated in Figure \ref{fig:1}, our framework is composed of two primary components: a Text Processor and a Pose Generator. The Text Processor first transforms the input HamNoSys notation into high-dimensional contextual embeddings that guide the animation process. Subsequently, the Pose Generator receives these embeddings and produces a corresponding sequence of human poses using an iterative, coarse-to-fine refinement strategy. The following sections detail each part of the architecture.

\subsection{Text Processor}
The Text Processor module is responsible for converting the input HamNoSys sequence into a sequence of contextual embeddings that guide the pose generator. Our architecture adopts our baseline Ham2Pose model text processing pipeline \cite{arkushin2023ham2pose}.

To process Hamburg Notation System  (HamNoSys) symbol sequences into a format suitable for our model, we first tokenize the input sequence. The tokenization process can be formally described as follows:

Let $V = \{v_1, v_2, \ldots, v_n\}$ be the set of all HamNoSys symbols. In our implementation, each HamNoSys symbol $v_i$ corresponds to a single Unicode character.
The vocabulary $V$ is constructed by extracting the available glyph codepoints from the cmap table of the HamNoSys font file.
Therefore, $n = |V|$ denotes the number of valid HamNoSys Unicode characters (excluding special tokens). In our setting, $n=210$, and the embedding vocabulary size is $n+2=212$ when including \texttt{PAD}=0 and \texttt{BOS}=1.
We can define a Token Mapping function $f: V \rightarrow \{2, 3, \ldots, n+1\}$ such that each $v \in V$ is mapped to a unique integer id in this range.

The inverse function $f^{-1}$ maps token IDs back to symbols.
In addition, there are two special tokens: PAD = 0 (padding token) and BOS = 1 (beginning of sequence token).
So overall, for a HamNoSys sequence $S = (s_1, s_2, \ldots, s_k)$, where $s_i \in V$, we define the tokenization function $T$:
    \[T(S) = [BOS, f(s_1), f(s_2), \ldots, f(s_k)]. \]
    
For a batch of $m$ sequences $B = \{S_1, S_2, \ldots, S_m\}$, let $l = \max(|T(S_i)|)$ for $i \in \{1, 2, \ldots, m\}$. We define the padding function $P$:
    \[P(T(S), l) = [T(S) + [PAD] * (l - |T(S)|)] . \] 
    The batch tokenization function $TB$ is then:
    \[TB(B) = \{P(T(S_1), l), P(T(S_2), l), \ldots, P(T(S_m), l)\}. \] 

And to enhance Transformer's ability to handle variable-length sequences, the model includes the attention mask. For a padded sequence $PS = P(T(S), l)$, the attention mask $A$ is defined as:
    \[A(PS) = [1 \text{ if } t = PAD \text{ else } 0 \text{ for } t \in PS],\] where 1 (True) indicates padding positions that should be ignored by attention. 
For example, the input HamNoSys string \texttt{\textbackslash ue000\textbackslash ue071}, the token ids are
$[1, 11, 95, 0, 0, 0, 0, 0, 0, 0, 0, 0]$,
where 1 is \texttt{BOS} and 0 is \texttt{PAD}.
The corresponding padding mask is
$[\text{F}, \text{F}, \text{F}, \text{T}, \text{T}, \text{T}, \text{T}, \text{T}, \text{T}, \text{T}, \text{T}, \text{T}]$,
where \text{T} indicates padding positions.

To further enhance the Transformer's ability to capture the sequential information in HamNoSys sequences, the tokenized sequences are then passed to positional embedding layers. For a padded sequence $PS$ of length $l$, we define a learnable positional embedding function $PE$ as follows:
\[PE: {1, 2, ..., l} \rightarrow \mathbb{R}^d , \]
where $d$ is the hidden dimension of the model. For each position $i$,
\[PE(i) = \mathbf{e}_i \in \mathbb{R}^d .\] 
Here, $\mathbf{e}_i$ is a learnable vector that represents the embedding of position $i$.
The final input representation $I$ for a token $t$ at position $i$ in the padded sequence $PS$ is then given by:
\[I(t, i) = E(t) + PE(i) ,\] 
where $E(t)$ is the embedding of the token $t$. This approach allows the model to learn optimal representations for different positions in the sequence, rather than using predefined positional encoding functions. 

After being processed by the token embedding, these combined embeddings are then passed to the Transformer Encoder with attention mask to encode the text. Finally, the encoded text is passed to the sequence length predictor $\mathbb{R}^d \rightarrow \mathbb{R}^1$ and pose generator to generate the pose output.

The sequence length predictor, inherited from the baseline model Ham2Pose, processes the encoded text representations to predict the length of the output pose sequence, and it is supervised by ground truth sequence lengths during training. During inference, the Sequence Length Predictor is used to determine the appropriate length of the generated sign language motion sequence. This prediction process can be formally described as follows:

Let $H \in \mathbb{R}^{B \times N \times D}$ be the encoded text representation from the Transformer Encoder, where $B$ is the batch size, $N$ is the sequence length, and $D$ is the hidden dimension. We define a linear projection function $L$ that maps the hidden representations to scalar values:
$$L: \mathbb{R}^d \rightarrow \mathbb{R}^1. $$
For each token representation in the encoded sequence, we apply this projection:
$$l_{i,j} = L(h_{i,j}) \quad \text{for } h_{i,j} \in H. $$
The sequence length prediction for each sequence in the batch is obtained by averaging across the sequence dimension:
$$\hat{l}_i = \frac{1}{n}\sum_{j=1}^n l_{i,j}, $$
where $\hat{l}_i$ represents the predicted sequence length for the $i$-th sequence in the batch.

To ensure that the predicted sequence length falls within valid bounds, we apply a clipping function:
$$\hat{l}_{final} = \text{clip}(\hat{l}, l_{min}, l_{max}), $$
where $l_{min}$ and $l_{max}$ are the minimum and maximum allowed sequence lengths respectively.

\subsection{Pose Generator Overview}

To generate high-quality sequence outputs and ensure temporal coherence between consecutive frames, we adopt an approach based on our baseline \cite{arkushin2023ham2pose} that utilizes decreasing step sizes to iteratively refine the output. This method first generates coarse details and then progressively refines them. By ensuring that larger changes occur in earlier steps, the model can establish the overall pose structure before focusing on nuanced movements. This approach allows the model to generate sequences gradually, avoiding the instability that might arise from generating the entire sequence in a single step. Consequently, it enables the model to progressively correct errors or fill in missing information.

This generation process unfolds over a series of $T$ steps, with the initial sequence at step $T$ ($s_T$) being an extension of the provided reference frame to match the target sequence length. During the training phase, this length corresponds to the actual duration of the input sign video, while during inference, the sequence length is predicted by the sequence length predictor.
To achieve a gradual and refined generation of the desired sign, we define a schedule function $\gamma \in [0, 1]$ as:
\begin{equation}\gamma_t = \log_T(T - t + 1).\end{equation}
Building on this, we calculate a dynamic step size $\alpha_t$ for each time step:
\begin{equation}\alpha_t = \gamma_t - \gamma_{t+1}.\end{equation}
The core of the model generation process lies in the computation of the predicted pose sequence $\hat{s}_t$ at each time step $t$, for $t \in {T-1, \ldots, 0}$:
\begin{equation}
\hat{s}_t = \alpha_t q_t + (1 - \alpha_t) \hat{s_{t+1}} ,
\end{equation}
where $q_t$ represents the pose value output by refinement module at time step $t$.

During training, we employ Teacher Forcing to stabilise the learning process. At each step, with probability $p_{tf}$, we replace the model's prediction $\hat{s}_t$ with the ground truth pose $s_t^{gt}$:
\begin{equation}
s_t = \begin{cases} 
   s_t^{gt} & \text{with probability } p_{tf} \\
   \hat{s}_t & \text{with probability } 1 - p_{tf} .
\end{cases}
\end{equation}

As the generation progresses and $t$ decreases, the step size $\alpha_t$ becomes smaller, allowing for increasingly fine-grained adjustments. So the overall generation process initially focuses on broader, coarser details before refining the finer aspects of the pose sequence. And by blending the current prediction with the previous step's result, the pose generator progressively reduces the influence of the initial pose sequence, which enables the model to address and correct missing or inaccurate keypoints through gradual refinement. The smooth, incremental nature of the process leads to more natural and accurate pose sequences compared to methods that attempt to predict the entire sequence in one step.
To enhance the robustness of the model during training, we incorporate noise $\epsilon z$ (where $z \sim \mathcal{N}(0, I)$) to $\hat{s}_t$ at each time step. 
The final output of our pose generator is $s_0$, representing the culmination of all refinement steps. 

\subsection{Text Pose Encoder}

The Text Pose Encoder is responsible for encoding the initial pose sequence, integrating it with HamNoSys text embeddings, and learning a suitable pose representation through iterative refinement.

The model processes several inputs, including a sequence of poses represented as a 4-dimensional tensor $X \in \mathbb{R}^{B \times T \times K \times 2}$. In this tensor, $B$ denotes the batch size, $T$ is the number of frames in the sequence, and $K$ is the number of keypoints, set to 137 in our work. The final dimension corresponds to the x and y coordinates of each keypoint.

Building upon the input tensor $X$, the pose sequence is first initialized by replicating the first frame $x_0$ to match the target sequence length T:
\begin{equation}
X_{init} = [x_0, x_0, ..., x_0] \in \mathbb{R}^{B \times T \times K \times 2} ,
\end{equation}
where $B$ is the batch size, $T$ is the sequence length determined by the ground truth during training or the sequence length predictor during inference, K is the number of keypoints, and 2 represents the x,y coordinates.

The pose embedding process then proceeds as follows:

First, the initial pose sequence $X_{init} \in \mathbb{R}^{B \times T \times K \times 2}$ is flattened to $X_{flat} \in \mathbb{R}^{B \times T \times 2K}$, which simplifies subsequent operations by combining the keypoint and coordinate dimensions.

A linear projection module, denoted by \( f_p : \mathbb{R}^{2K} \rightarrow \mathbb{R}^{D} \), is applied to map the pose representation into the Transformer hidden space, where \(D\) is the hidden dimension.
\begin{equation}
X_{proj} = f_p(X_{flat}) .
\end{equation}
Positional embeddings are generated using a learnable function $E_{pos}: \mathbb{N} \rightarrow \mathbb{R}^D$:
\begin{equation}
P = E_{pos}(t), \quad t \in \{0, 1, ..., T-1\} .
\end{equation}
The final pose embedding is computed by adding the projected pose and positional embedding, allowing the model to capture both spatial and temporal features:
\begin{equation}
E_p = X_{proj} + P .
\end{equation}
Concurrently, the step encoder generates a unique encoding for each refinement step, allowing the model to adapt its behavior based on the current stage of the generation process. It is implemented as a series of linear transformations with non-linear activations:
\begin{equation}
E_s = f_2(\text{SiLU}(f_1(e(t)))) ,
\end{equation}
where $e(t)$ is an embedding lookup for the current step $t$, $f_1$ and $f_2$ are linear transformations, and SiLU is the Sigmoid Linear Unit activation function. The resulting step encoding $E_s \in \mathbb{R}^{B \times 1 \times D}$ provides temporal context for the refinement process.

Finally, the pose embedding $E_p$ is concatenated with the text encoding $E_t \in \mathbb{R}^{B \times L \times D}$ (where $L$ is the text sequence length) and the step encoding $E_s$ to form a comprehensive representation $S = [E_p; E_t; E_s]$. The concatenated sequence is processed through a transformer encoder called text-pose encoder:
\begin{equation}
E_{text\_pose} = \text{TransformerEncoder}(S) .
\end{equation}
The output of the text-pose encoder $E_{text\_pose}$, is then passed to the Coarse Pose Module for further processing.

\subsection{Multi-Scale Skeleton Pose Animation}

\subsubsection{Coarse Pose Module}

The coarse module is a key component of our improved pose generator, designed to strengthen the model’s understanding of global posture and joint dependencies. Baseline models typically treat all joints equally, overlooking the hierarchical nature of the human body and the structural connections between joints. To address this, the coarse module predicts a simplified representation of the pose from the text–pose encoding, introducing global structural supervision into the generation process.

Because the dataset does not provide coarse-level annotations, we derive them from the ground truth fine poses. Specifically, the 137 keypoints are grouped into 25 body parts covering the main arm segments, the head, and the proximal and distal components of each finger on both hands. For each part, the mean position of its keypoints is computed, producing a compact coarse pose that preserves essential body structure while reducing local detail. This representation captures the overall skeletal organization and facilitates more consistent modeling of arm articulation and hand configurations, which are critical for accurate sign language generation.

\begin{figure}[!tbp]
\centering
\includegraphics[width=0.75\columnwidth]{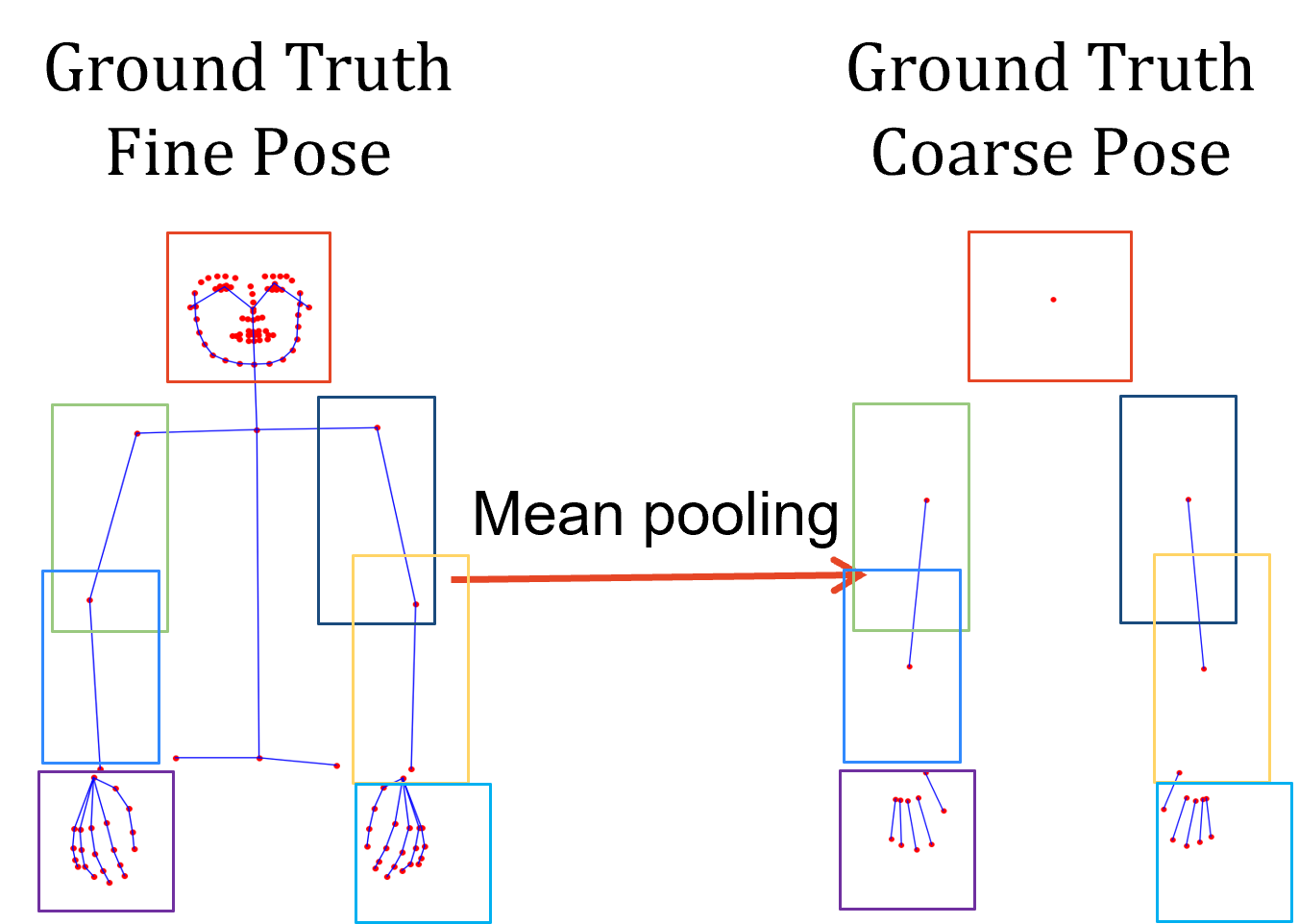}
\caption{We divide the keypoints from the original dataset into 25 body parts, including major segments of the arms, head, and detailed finger joints for both hands.}
\label{fig:coarse_pose_extraction}
\end{figure}

\textbf{Ground Truth Coarse Pose Extraction.} For each of the 25 predefined body key-parts, we first retrieve the 2D coordinates of all keypoints corresponding to the indices specified for that body part. We then calculate the mean of these keypoints' coordinates. The resulting mean positions collectively form our coarse pose representation. An example illustration of multi-scale skeleton pose extraction is shown in Figure \ref{fig:coarse_pose_extraction}. This process can be formalized as follows.

For each body part $p$ in the 25 body key-parts:
\begin{equation}
\text{CoarsePose}_{gt}[p] = \frac{1}{|I_p|} \sum{i \in I_p} \text{Pose}_i ,
\end{equation}
where $I_p$ is the set of indices for body part $p$, and $\text{Pose}_i$ is the 2D coordinate of the $i$-th keypoint in the original pose.

Thus, the complete ground truth coarse pose $\text{CoarsePose}_{gt}$ is a tensor of shape $[B, T, 25, 2]$, where $B$ is the batch size, $T$ is the number of frames, 25 is the number of key body parts, and 2 represents the x and y coordinates.

\textbf{Coarse Pose Prediction.} After extracting the ground truth coarse pose, we need the model to learn how to generate coarse poses from the encodings produced by the Text Pose Encoder. The coarse pose projection network is designed to generate a coarse pose from the text pose encoding, so that we can capture the essential global structure of the human body. This prediction process is supervised by the ground-truth coarse poses extracted in the previous step. The projection is implemented as follows:
\begin{equation}
\text{CoarsePose}_{pred} = f_2(\text{SiLU}(f_1(E_{tp}))) ,
\end{equation}
where $f_1: \mathbb{R}^D \rightarrow \mathbb{R}^D$ and $f_2: \mathbb{R}^D \rightarrow \mathbb{R}^{50}$ are linear transformations, $D$ is the hidden dimension, and $E_{tp}$ is the output of the text-pose encoder. The output dimension 50 corresponds to the 25 key body parts, each represented by a 2D coordinate.

This approach allows the model to learn the mapping between fine pose and coarse pose representations, ensuring that the generated coarse pose accurately captures the overall structure and key features of the human pose. By incorporating this coarse pose prediction and supervision, the model gains a better understanding of the global posture and joint connections, which can lead to more coherent and anatomically plausible pose generations in the subsequent fine pose prediction stage.

\subsubsection{Fine Pose Prediction}

\begin{figure}[!tbp]
\centering
\includegraphics[width=0.90\columnwidth]{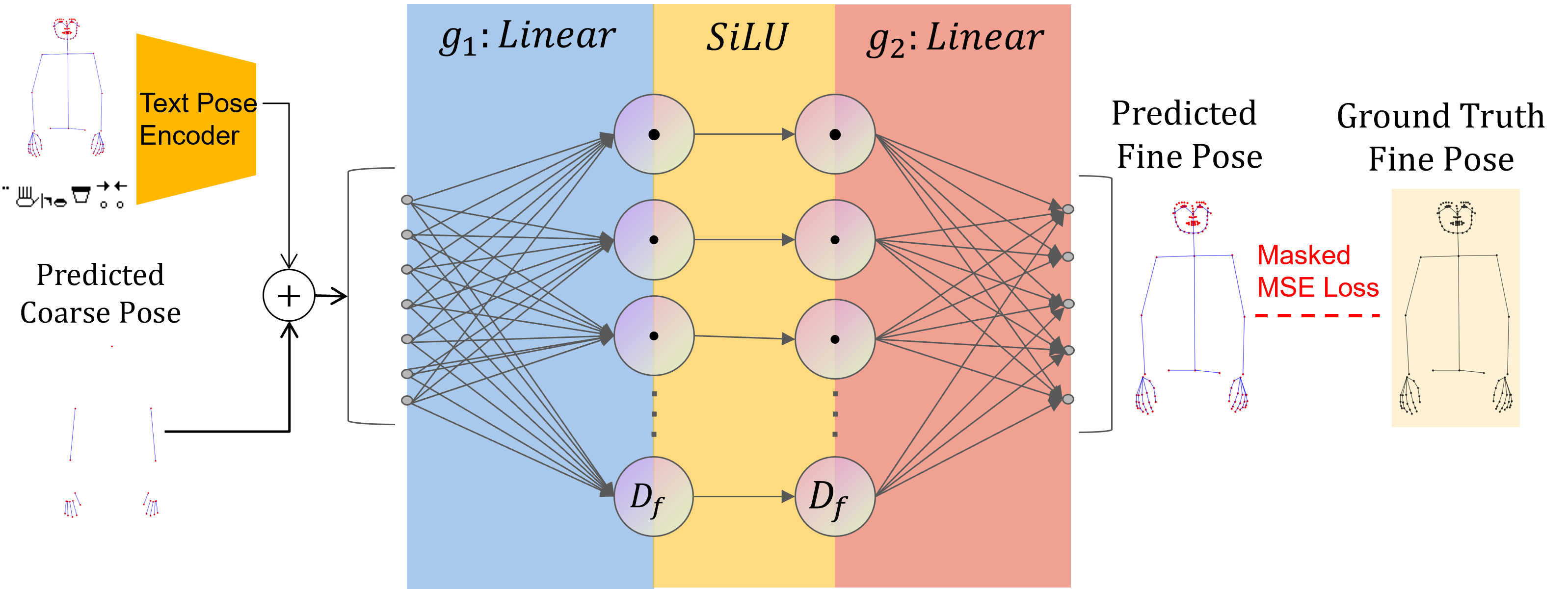}
\caption{Architecture of the fine pose prediction module.}
\label{fig:fine_pose_prediction}
\end{figure}

The fine pose prediction network (See Figure \ref{fig:fine_pose_prediction}) generates the final detailed pose by combining the encoded pose features from the text-pose encoder and the predicted coarse pose. This process can be formalized as follows.
\begin{equation}
\text{FinePose}_{\text{pred}} = g_2(\text{SiLU}(g_1(E_{tp} + \text{CoarsePose}_{\text{pred}}))).
\end{equation}
where $g_1: \mathbb{R}^{D + 50} \rightarrow \mathbb{R}^D$ and $g_2: \mathbb{R}^D \rightarrow \mathbb{R}^{274}$ are linear transformations. $D$ is the hidden dimension, 50 represents the dimension of the coarse pose (25 key body parts × 2 coordinates), and 274 is the dimension of the final pose (137 keypoints × 2 coordinates).
The concatenation of $E_{tp}$ (encoded pose features) and $\text{CoarsePose}_{\text{pred}}$ (predicted coarse pose) allows the model to leverage both global structural information and detailed feature representations to produce the final pose. This approach enables the network to refine the coarse pose prediction while maintaining global coherence and incorporating fine pose details.

By combining the coarse and fine pose predictions in this manner, KANMultiSign can generate detailed, anatomically plausible poses that maintain both global structure and local refinement.

\begin{figure*}[!t]
\centering
\resizebox{0.8\textwidth}{!}{
\begin{tabular}{c | c c c c c c }
\toprule
Frame & 0 & 10 & 20 & 30 & 40 & 50 \\
\hline
Fine pose & 
\includegraphics[width=2cm, height=3.62cm]{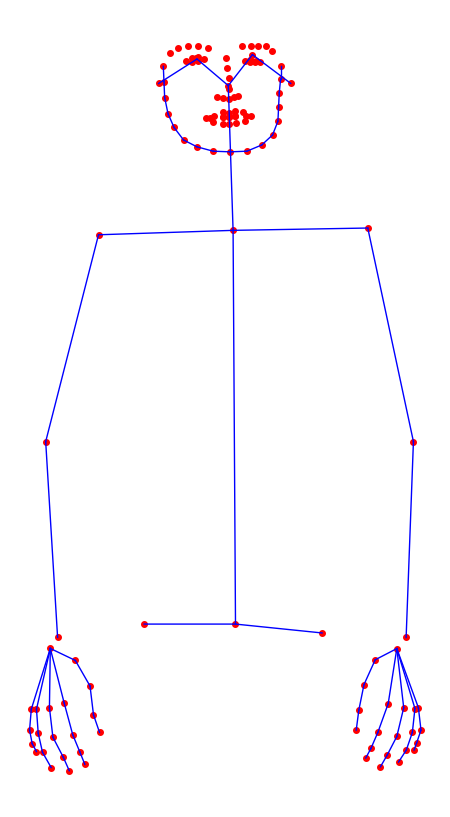} &
\includegraphics[width=2cm, height=3.62cm]{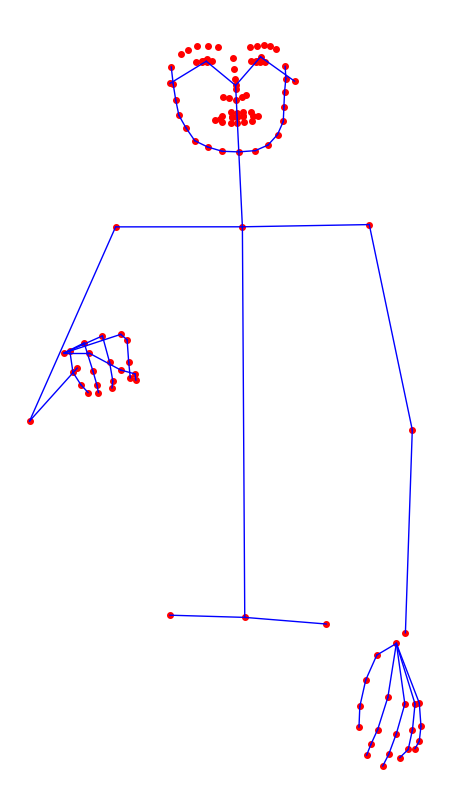} &
\includegraphics[width=2cm, height=3.62cm]{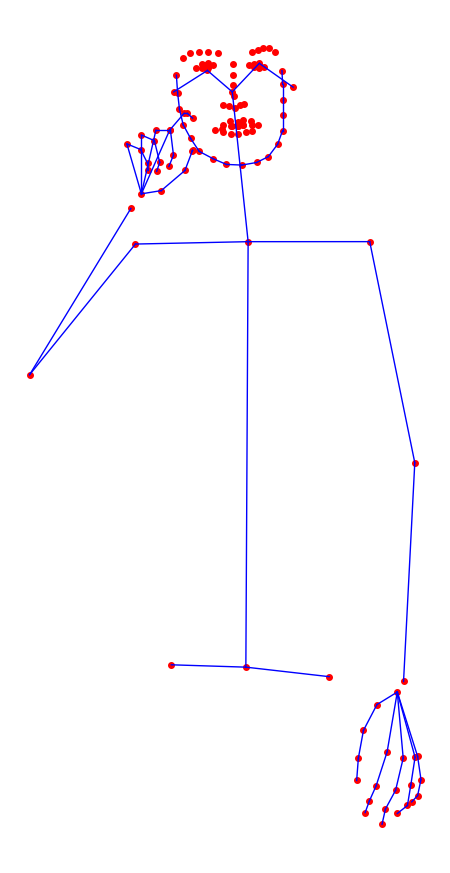} &
\includegraphics[width=2cm, height=3.62cm]{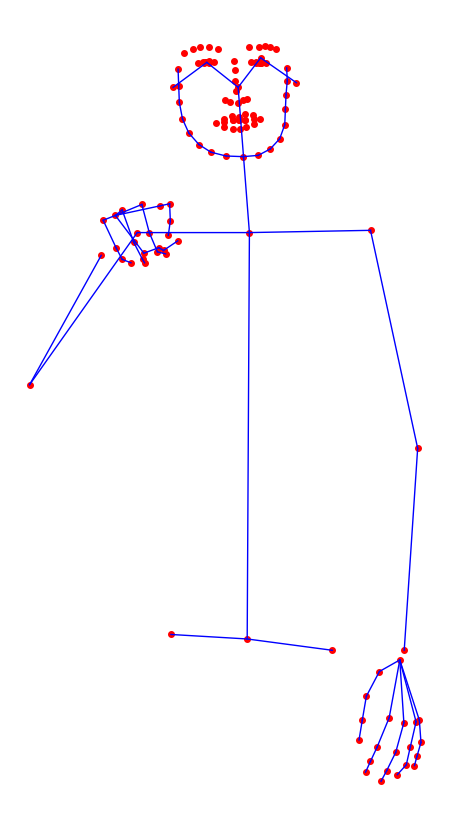} &
\includegraphics[width=2cm, height=3.62cm]{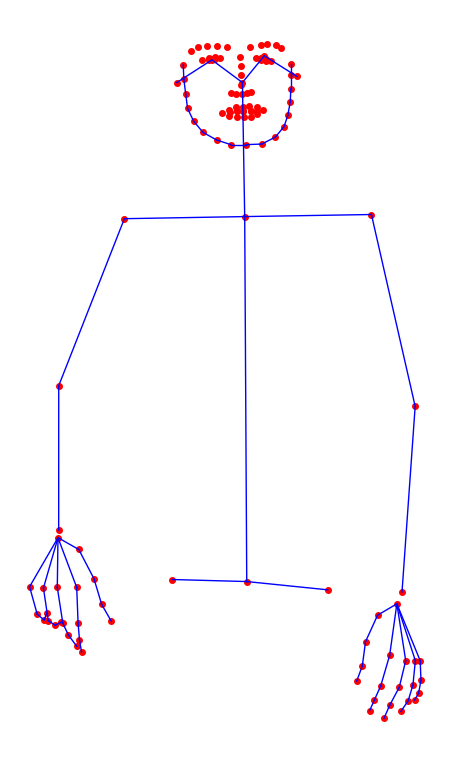} &
\includegraphics[width=2cm, height=3.62cm]{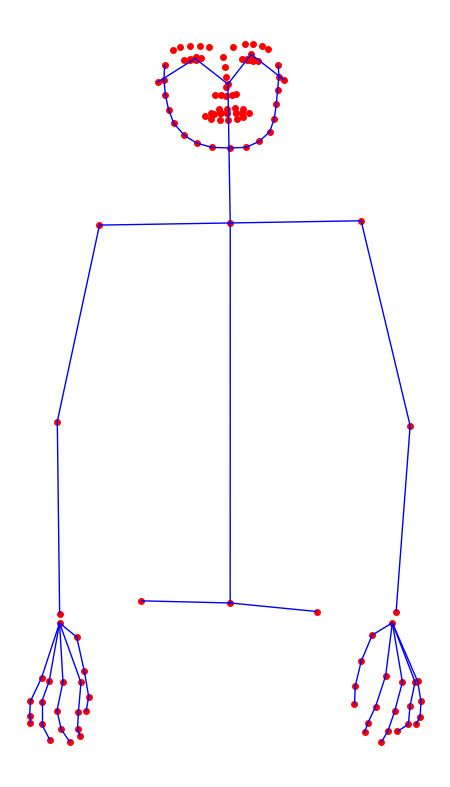} \\
\hline
Coarse pose & 
\includegraphics[width=2cm, height=3.62cm]{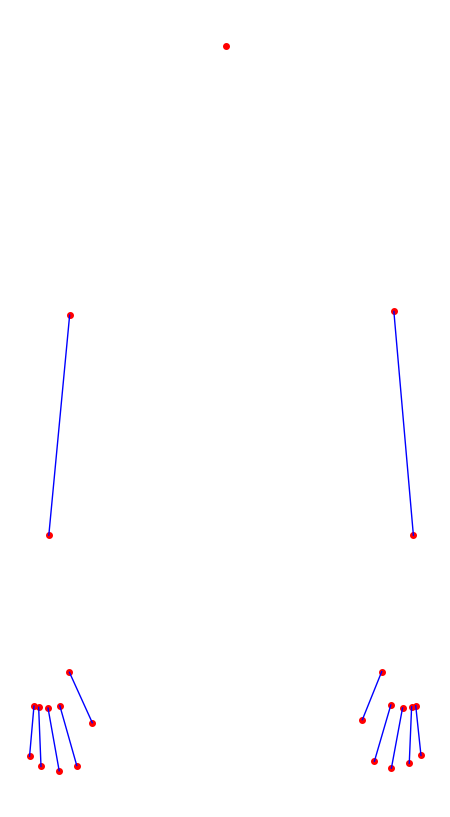} &
\includegraphics[width=2cm, height=3.62cm]{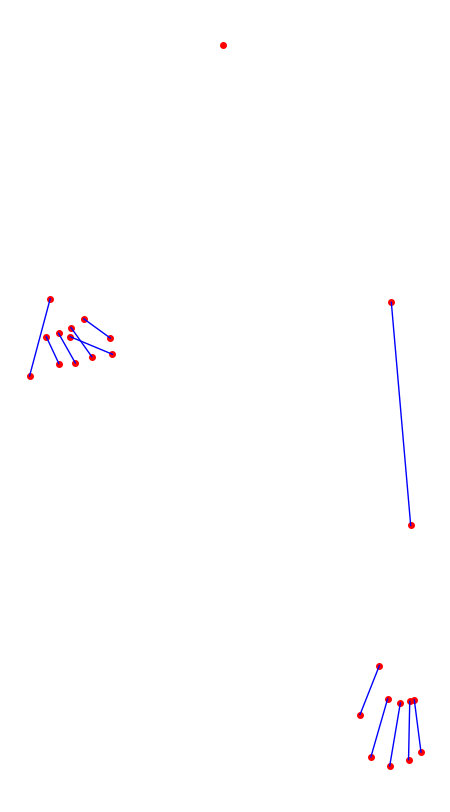} &
\includegraphics[width=2cm, height=3.62cm]{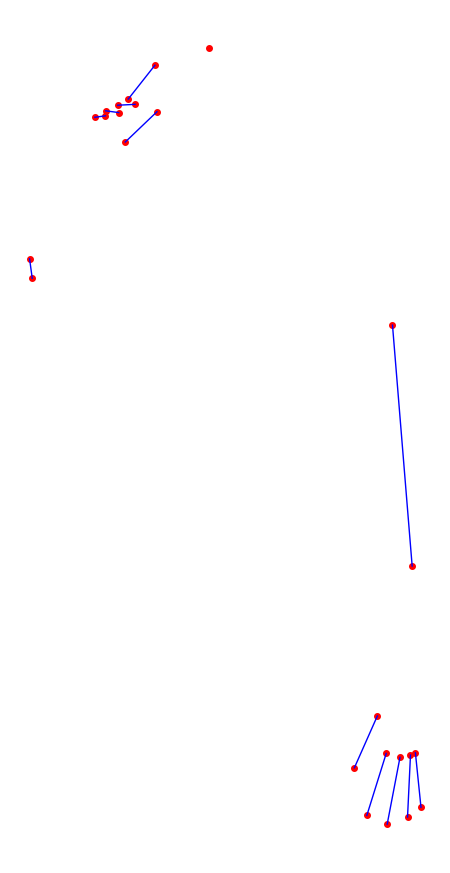} &
\includegraphics[width=2cm, height=3.62cm]{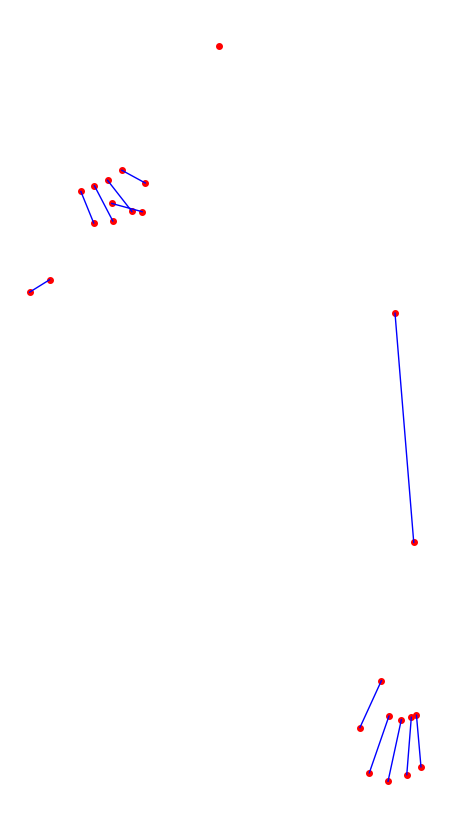} &
\includegraphics[width=2cm, height=3.62cm]{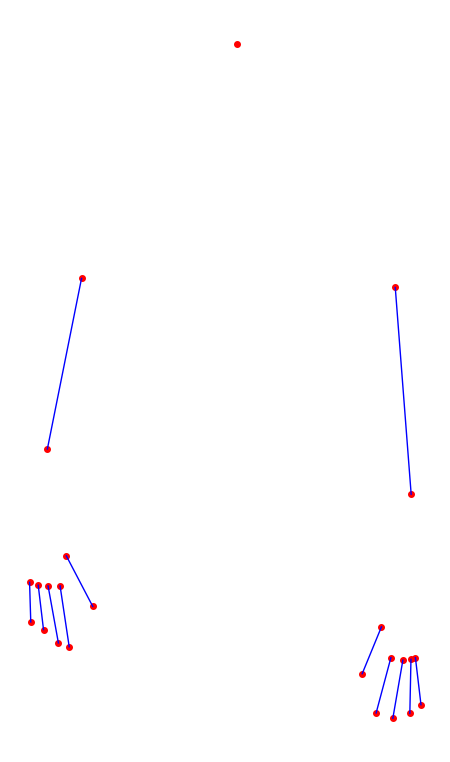} &
\includegraphics[width=2cm, height=3.62cm]{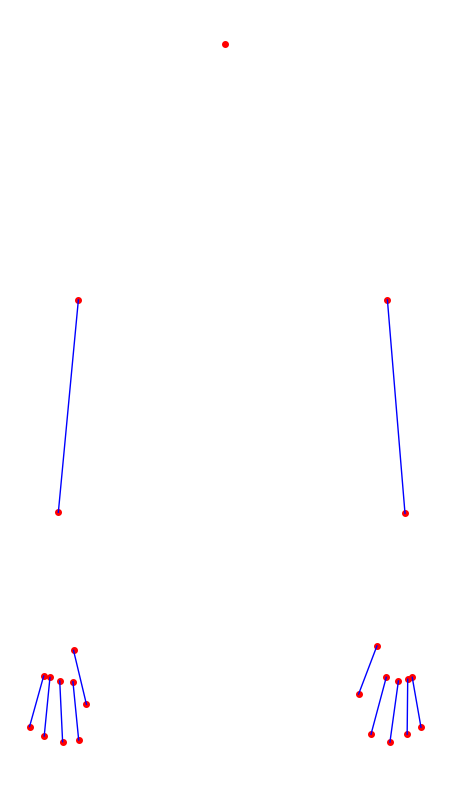} \\
\bottomrule
\end{tabular}
}
\caption{The comparison of fine and coarse poses at different frames.}
\label{tab:pose_comparison}
\end{figure*}

\subsection{KAN-based Transformer for Generation}

To enhance the parameter efficiency and modeling capacity of the Transformer's feed-forward layers, we replace them with KAN modules. Below, KAN will be briefly introduced.

An $L$-layer MLP can be written as interleaving of transformations ${W}$ and  activations $\sigma$:
\begin{equation}
\operatorname{MLP}(\mathbf{x})=\left({W}_{L-1} \circ \sigma \circ {W}_{L-2} \circ \sigma \circ \cdots \circ {W}_{1} \circ \sigma \circ {W}_{0}\right) \mathbf{x},
\end{equation}
which approximates complex functional mappings through multiple layers of nonlinear transformations. However, its deeply opaque nature constrains the model's interpretability, posing challenges to intuitively understanding the internal decision-making process.

To address the issues of low parameter efficiency and poor interpretability in MLPs, Liu \emph{et al.}~\cite{liu2024kan} introduced the KAN that is inspired by Kolmogorov-Arnold representation theorem ~\cite{kolmogorov1961representation,braun2009constructive}. 
Similar to MLP, a $L$-layer KAN can be described as a nesting of multiple KAN layers:
\begin{equation}
    \operatorname{KAN}(\mathbf{x})=\left(\boldsymbol{\Phi}_{L-1} \circ \boldsymbol{\Phi}_{L-2} \circ \cdots \circ \boldsymbol{\Phi}_{1} \circ \boldsymbol{\Phi}_{0}\right) \mathbf{x},
\end{equation}
where $\boldsymbol{\Phi}_i$ represents the $i$-th layer of the whole KAN network. For each KAN layer with $n_{in}$ -dimensional input and $n_{out}$ -dimensional output,  $\boldsymbol{\Phi}$ consists of $ n_{in} * n_{out}$ 1-D learnable activation functions $\phi$:
\begin{equation}
    \boldsymbol{\Phi}=\left\{\phi_{q, p}\right\}, \quad p=1,2, \cdots, n_{\text {in }}, \quad q=1,2 \cdots, n_{\text {out }}.
\end{equation}

When computing the result of the KAN network from layer $l$ to layer $l+1$, it can be represented in matrix form as follows:
\begin{equation}
    \mathbf{x}_{l+1}=\underbrace{\left(\begin{array}{cccc}
\phi_{l, 1,1}(\cdot) & \phi_{l, 1,2}(\cdot) & \cdots & \phi_{l, 1, n_{l}}(\cdot) \\
\phi_{l, 2,1}(\cdot) & \phi_{l, 2,2}(\cdot) & \cdots & \phi_{l, 2, n_{l}}(\cdot) \\
\vdots & \vdots & & \vdots \\
\phi_{l, n_{l+1}, 1}(\cdot) & \phi_{l, n_{l+1}, 2}(\cdot) & \cdots & \phi_{l, n_{l+1}, n_{l}}(\cdot)
\end{array}\right)}_{\boldsymbol{\Phi}_{l}} \mathbf{x}_{l}.
\end{equation}

In our generation backbone, we preserve the standard Transformer encoder structure (multi-head self-attention, residual connections, and layer normalization) and replace only the feed-forward network (FFN) with a KAN module.
Let $X \in \mathbb{R}^{B \times T \times d}$ denote the token representations (batch size $B$, sequence length $T$, model dimension $d$).
As in a standard Transformer, the FFN is applied independently to each token.
Accordingly, we define the KAN-based FFN in a token-wise manner as
\begin{equation}
\begin{aligned}
\mathrm{KAN\text{-}FFN}(X)_{b,t} &= \mathrm{Dropout}\big(\mathrm{KAN}(X_{b,t})\big), \\
&\qquad b=1,\dots,B,\; t=1,\dots,T.
\end{aligned}
\end{equation}

This formulation keeps temporal modeling solely within self-attention, while the KAN-FFN serves as a compact nonlinear mapping at the token level.

We parameterize the KAN-FFN with a three-layer width profile $[d, h, d]$, where $h$ is the intermediate width and serves as the primary capacity knob of the KAN-FFN under a fixed KAN basis configuration.
In all experiments, we keep the internal KAN basis/grid-related settings unchanged and only vary $h$ in the ablation study to analyze the accuracy--compactness trade-off.

\begin{figure*}[t]
    \centering
    \begin{subfigure}{\textwidth}
        \centering
        \includegraphics[width=0.09\textwidth]{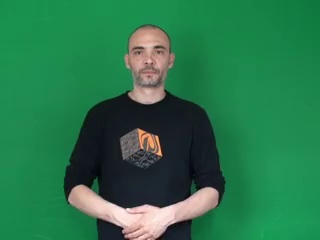}
        \includegraphics[width=0.09\textwidth]{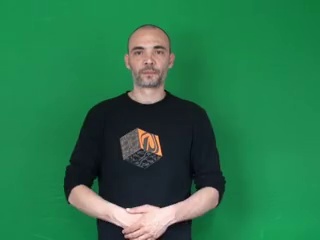}
        \includegraphics[width=0.09\textwidth]{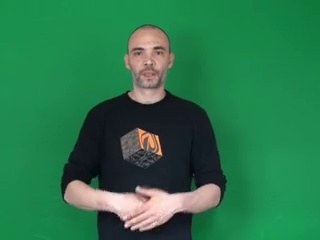}
        \includegraphics[width=0.09\textwidth]{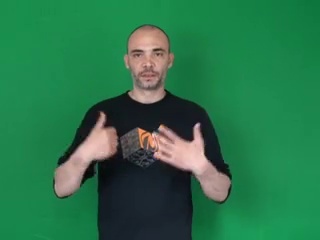}
        \includegraphics[width=0.09\textwidth]{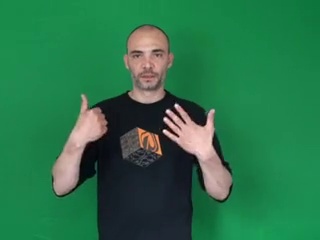}
        \includegraphics[width=0.09\textwidth]{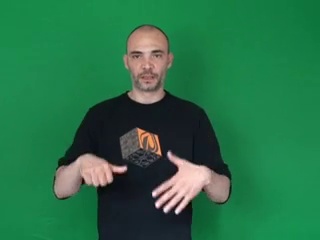}
        \includegraphics[width=0.09\textwidth]{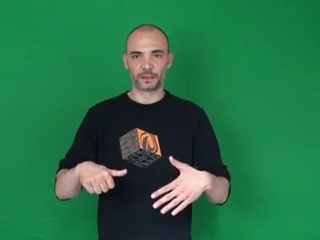}
        \includegraphics[width=0.09\textwidth]{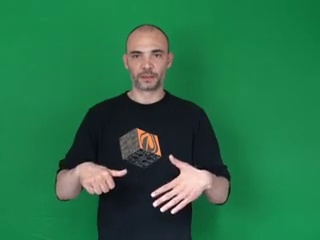}
        \includegraphics[width=0.09\textwidth]{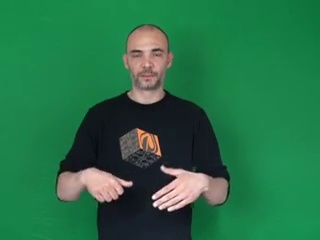}
        \includegraphics[width=0.09\textwidth]{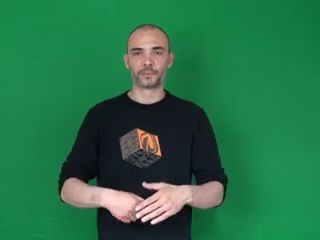}
        \caption{French Sign Language (LSF)}
    \end{subfigure}
    \begin{subfigure}{\textwidth}
        \centering
        \includegraphics[width=0.09\textwidth]{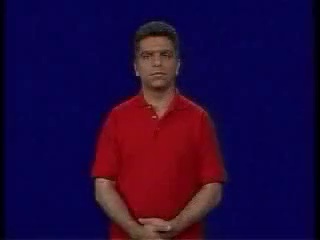}
        \includegraphics[width=0.09\textwidth]{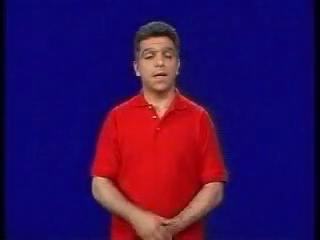}
        \includegraphics[width=0.09\textwidth]{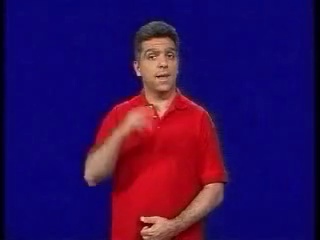}
        \includegraphics[width=0.09\textwidth]{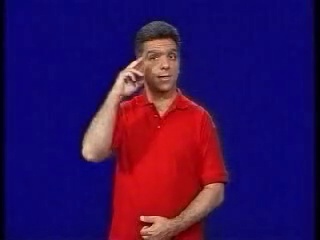}
        \includegraphics[width=0.09\textwidth]{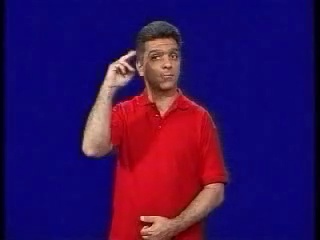}
        \includegraphics[width=0.09\textwidth]{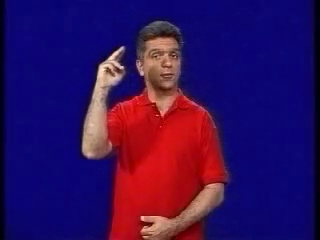}
        \includegraphics[width=0.09\textwidth]{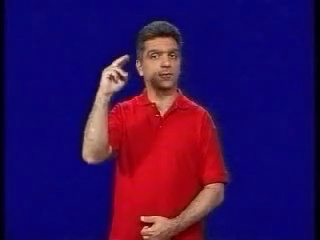}
        \includegraphics[width=0.09\textwidth]{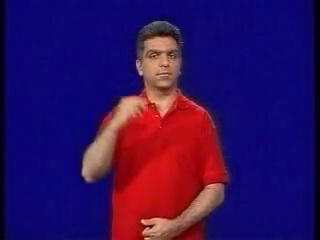}
        \includegraphics[width=0.09\textwidth]{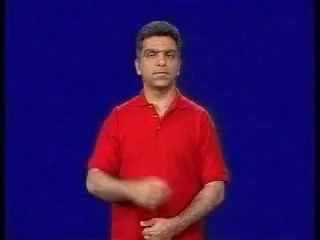}
        \includegraphics[width=0.09\textwidth]{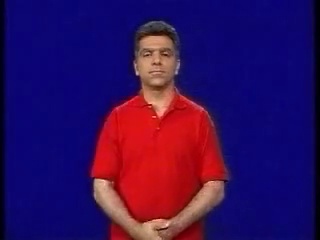}
        \caption{Greek Sign Language (GSL)}
    \end{subfigure}
    \begin{subfigure}{\textwidth}
        \centering
        \includegraphics[width=0.09\textwidth]{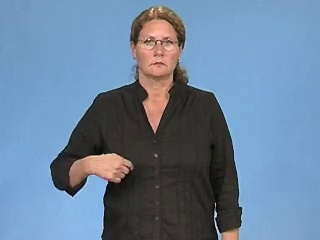}
        \includegraphics[width=0.09\textwidth]{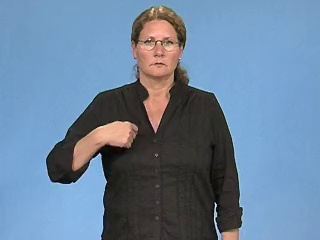}
        \includegraphics[width=0.09\textwidth]{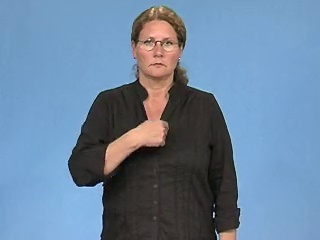}
        \includegraphics[width=0.09\textwidth]{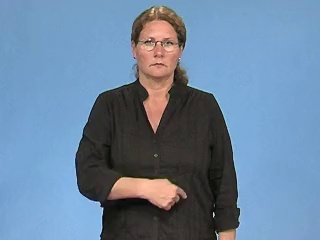}
        \includegraphics[width=0.09\textwidth]{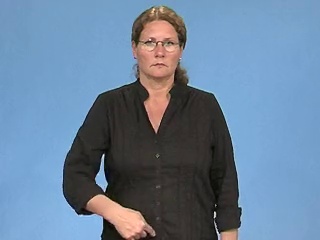}
        \includegraphics[width=0.09\textwidth]{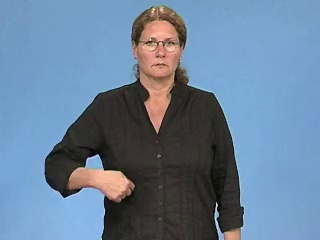}
        \includegraphics[width=0.09\textwidth]{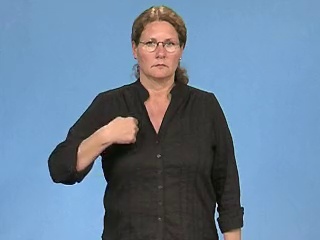}
        \includegraphics[width=0.09\textwidth]{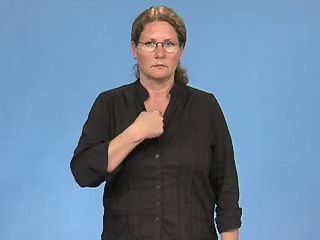}
        \includegraphics[width=0.09\textwidth]{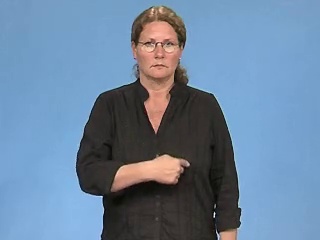}
       \includegraphics[width=0.09\textwidth]{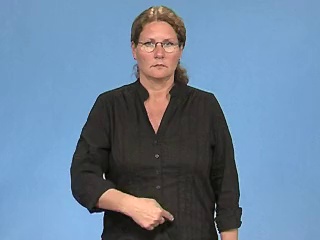}
        \caption{German Sign Language (DGS)}
    \end{subfigure}
    \begin{subfigure}{\textwidth}
        \centering
        \includegraphics[width=0.09\textwidth]{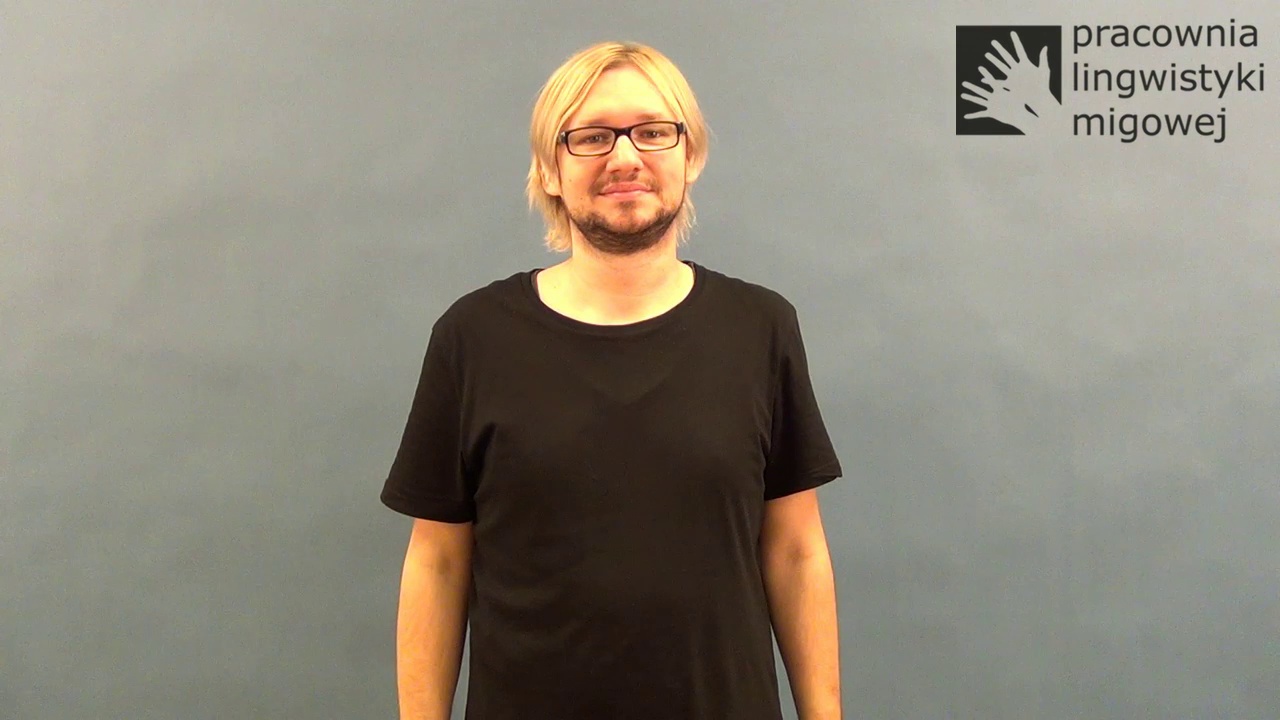}
        \includegraphics[width=0.09\textwidth]{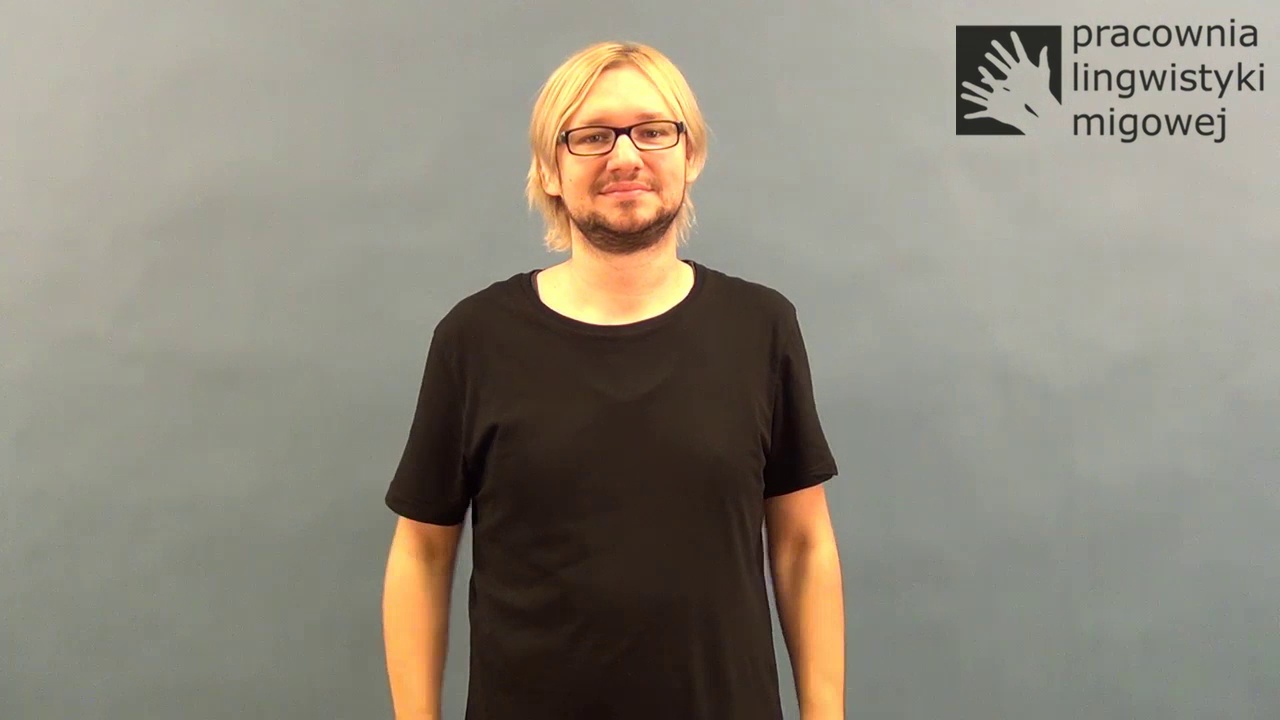}
        \includegraphics[width=0.09\textwidth]{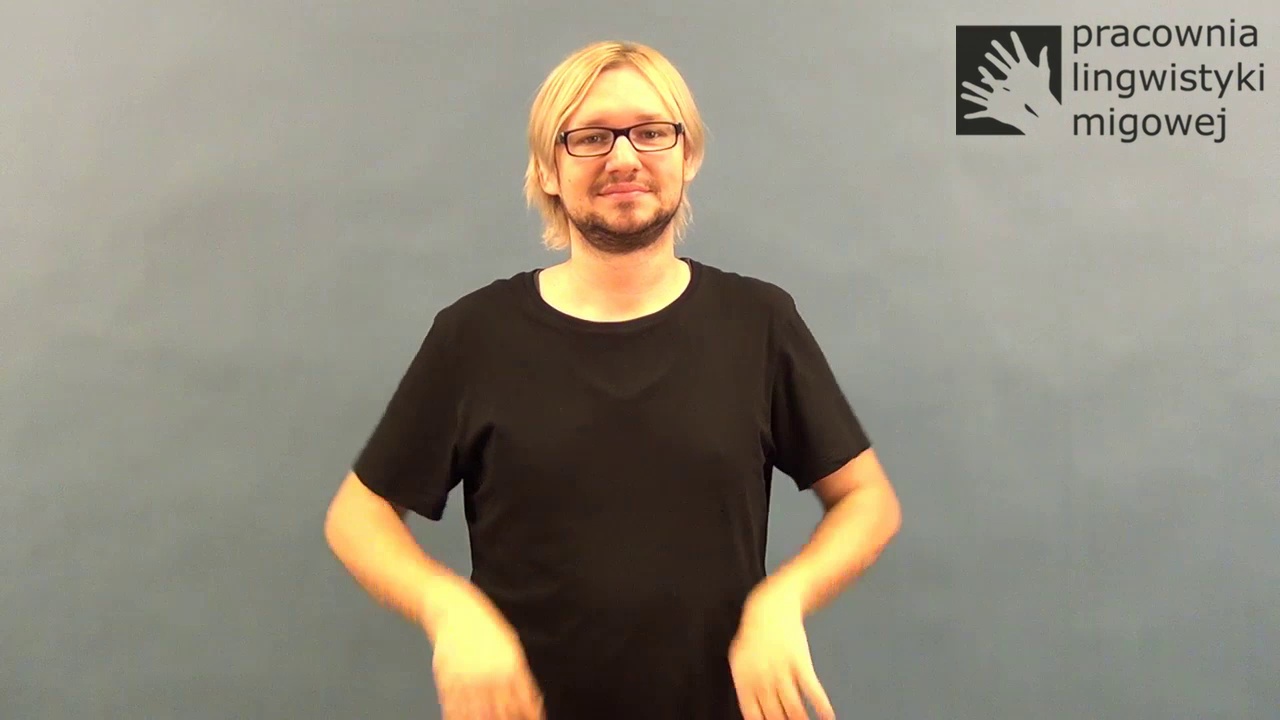}
        \includegraphics[width=0.09\textwidth]{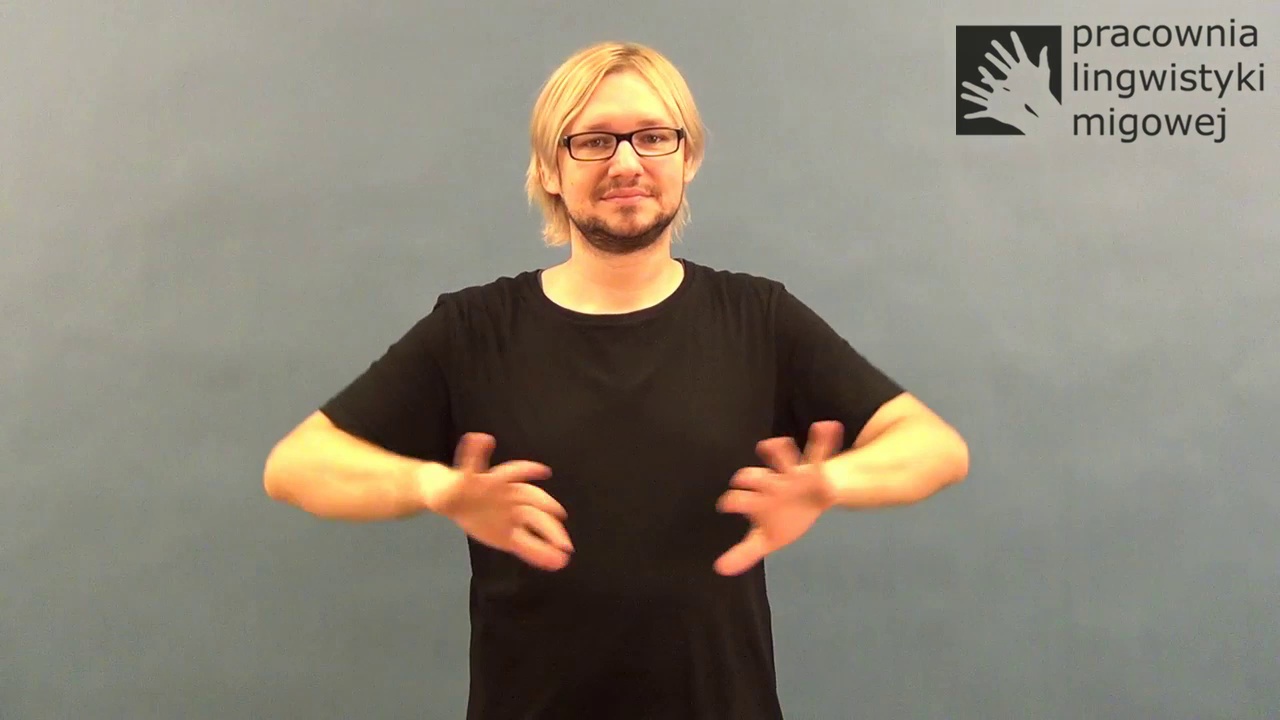}
        \includegraphics[width=0.09\textwidth]{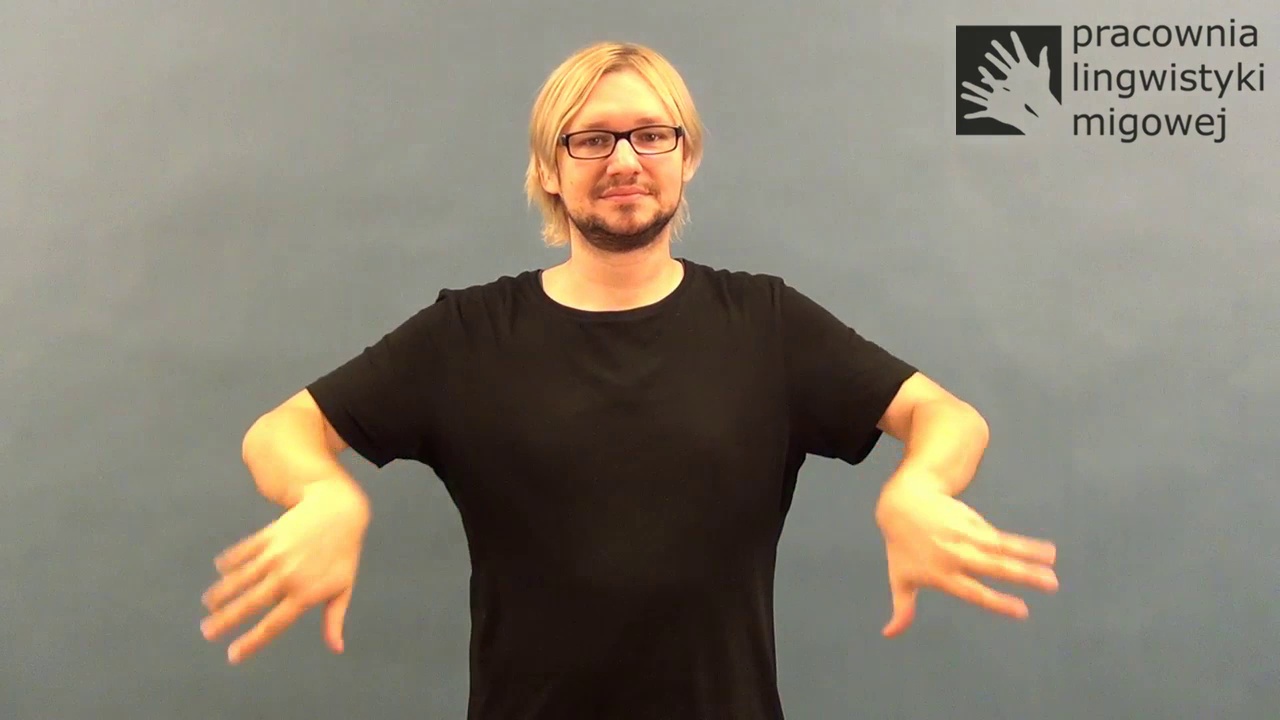}
        \includegraphics[width=0.09\textwidth]{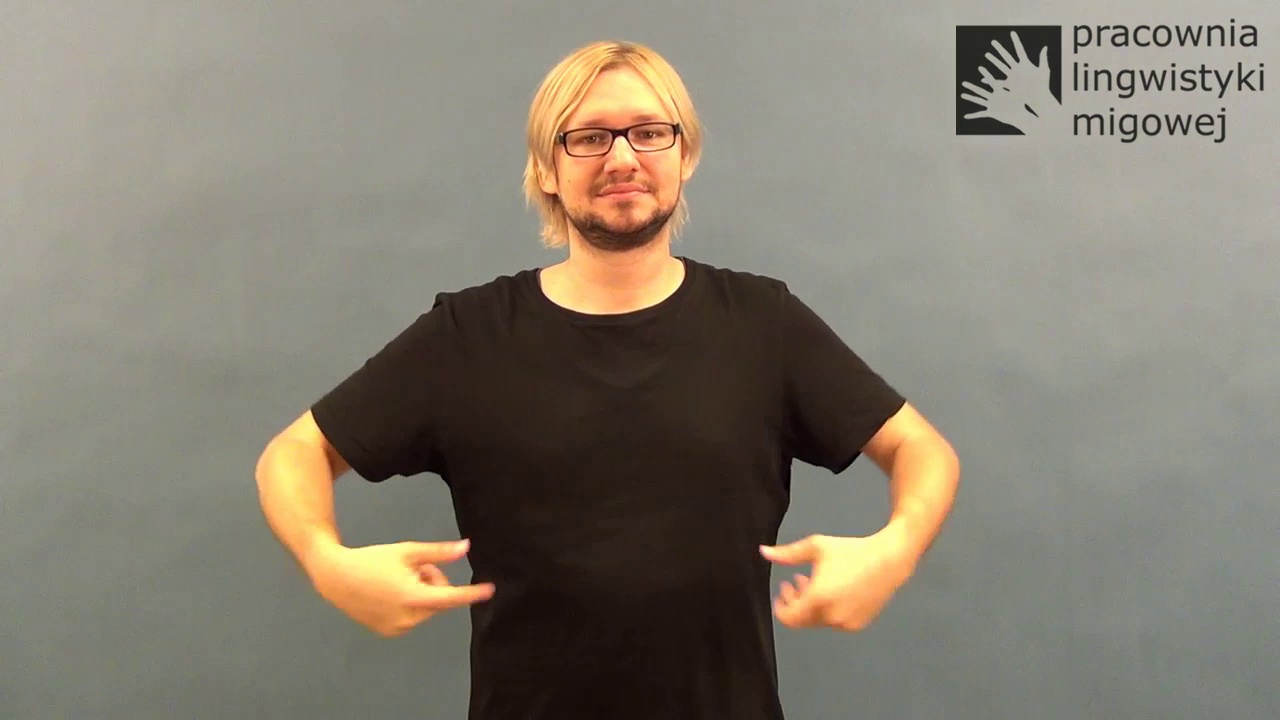}
        \includegraphics[width=0.09\textwidth]{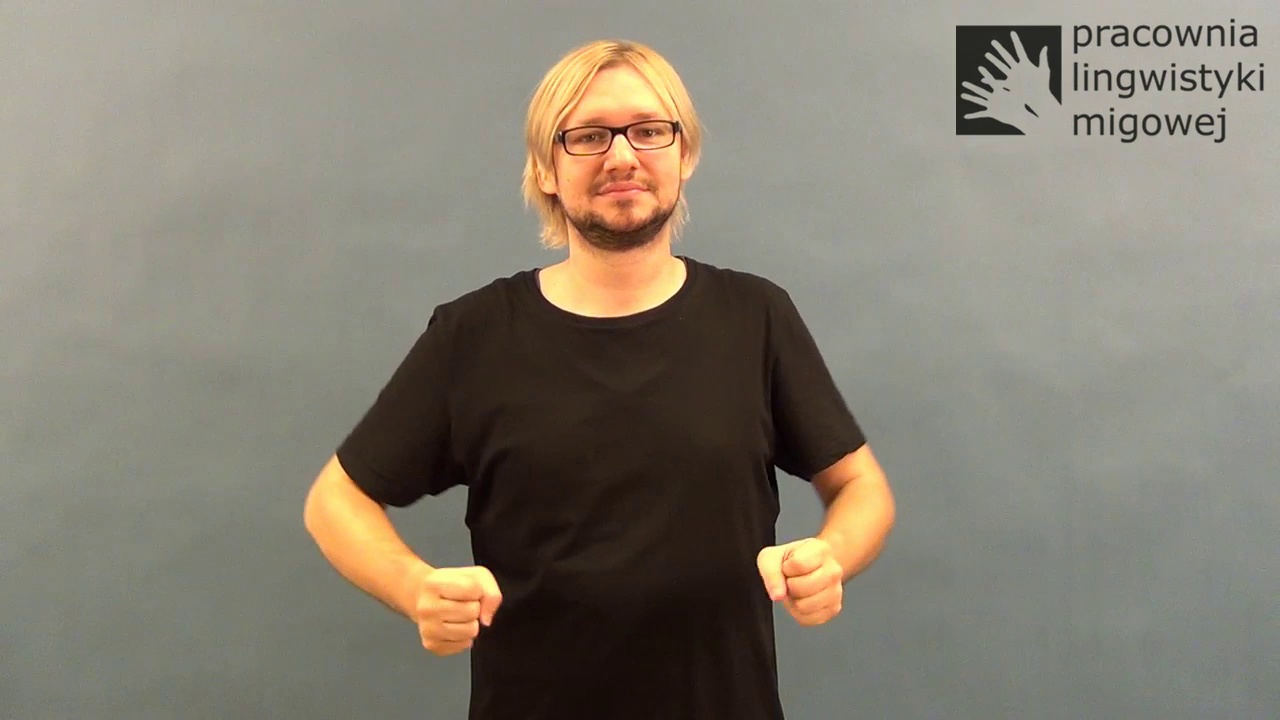}
        \includegraphics[width=0.09\textwidth]{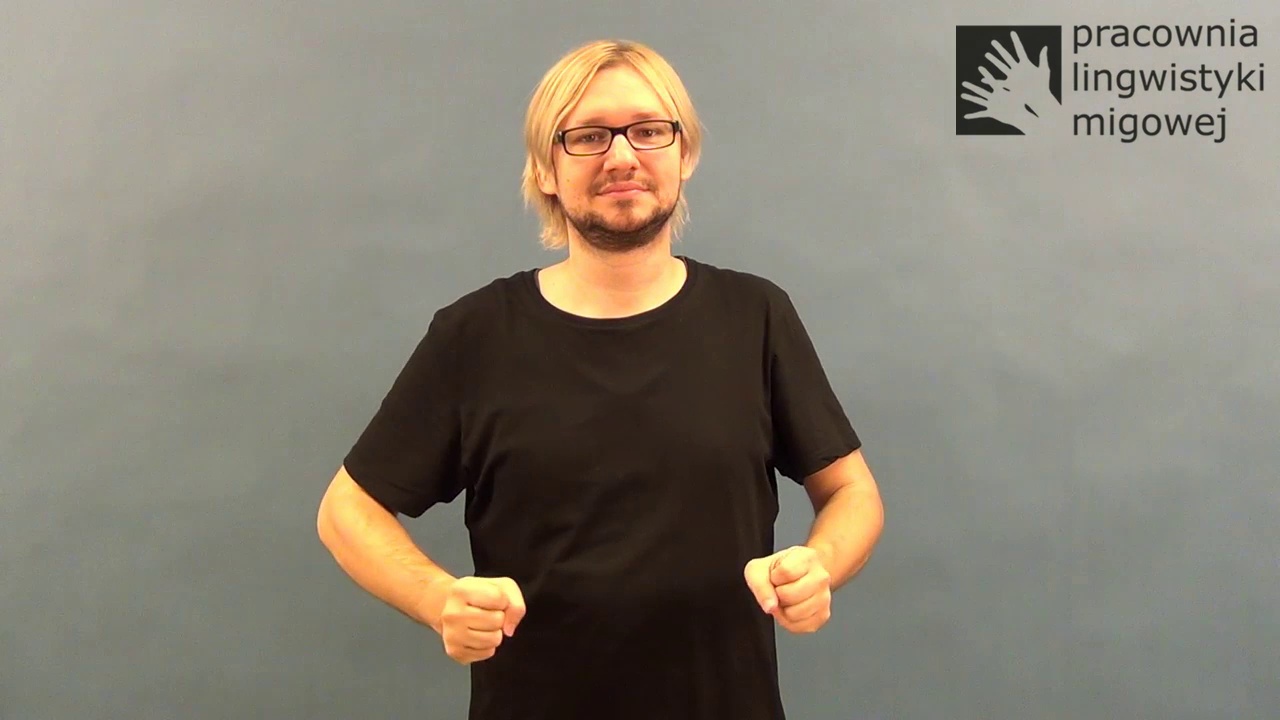}
        \includegraphics[width=0.09\textwidth]{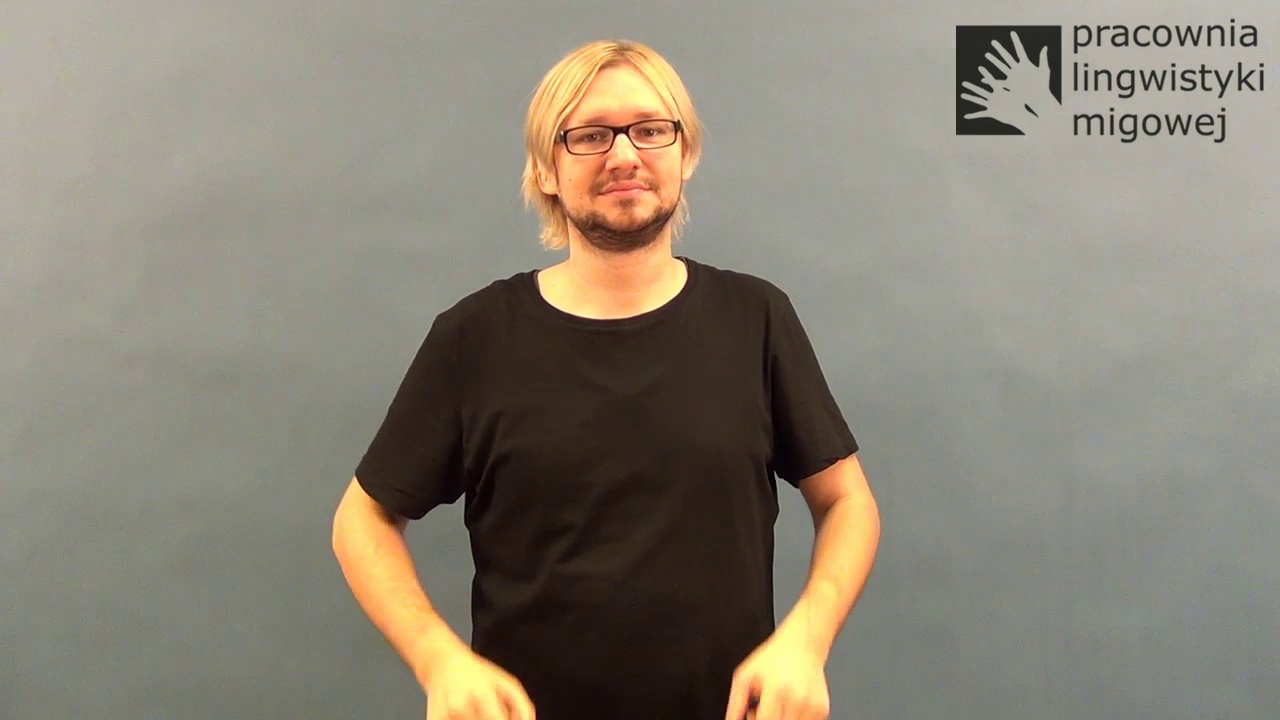}
       \includegraphics[width=0.09\textwidth]{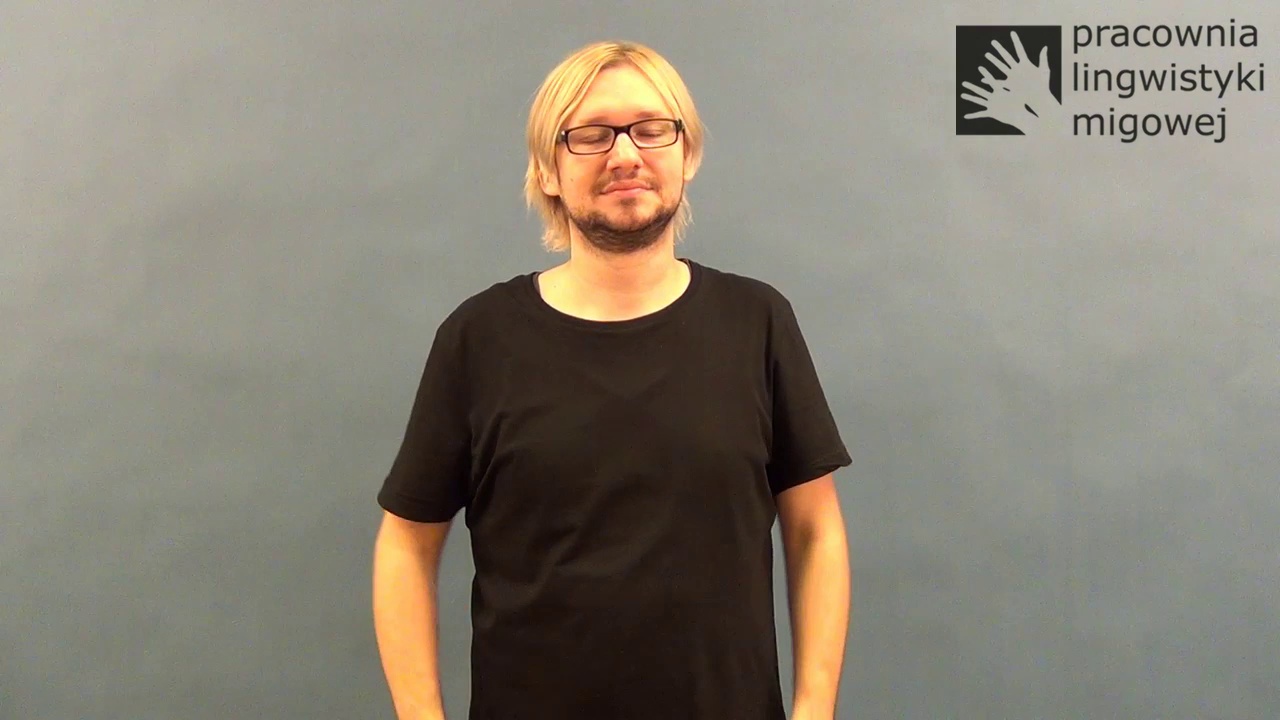}
        \caption{Polish Sign Language (PJM)}
    \end{subfigure}
    \caption{The examples of sign language sequences video from different datasets. Each row shows key frames from a single sign in the respective sign language.}
    \label{fig:sign_datasets}
\end{figure*}

\subsection{Loss Function}
\label{sec:loss}

The training of KANMultiSign is guided by a composite loss function that combines three distinct components. 

The sequence length loss $L_{seq\_length}$ guides the model to predict the correct length of the pose sequence. It is calculated as the Mean Squared Error (MSE) between the predicted sequence length and the ground truth sequence length. The MSE is defined as:

\begin{equation}
\text{MSE} = \frac{1}{n}\sum_{i=1}^{n}(y_i - \hat{y}_i)^2 ,
\end{equation}

where $n$ is the number of samples, $y_i$ is the ground truth value, and $\hat{y}_i$ is the predicted value. Therefore, the sequence length loss is formulated as:

\begin{equation}
L_{len} = \lambda  \text{MSE}(SequenceLength_{pred}, SequenceLength_{gt}).
\end{equation}

This component ensures that the generated pose sequences align temporally with the input sign language, maintaining the appropriate duration and rhythm of the signs. The weight of this loss to the total loss is controlled by the hyperparameter $\lambda$. To ensure a controlled and directly comparable setting with the Ham2Pose baseline~\cite{arkushin2023ham2pose}, we fix $\lambda = 2\times 10^{-5}$ for all experiments and do not perform additional tuning.

 we predict the coarse pose from the encoded fine-pose representation, supervised by the ground-truth coarse pose.
To enforce global postural consistency and guide the overall body structure, we introduce a coarse loss $L_{coarse}$. For a model with $T$ refinement steps,  we predict the coarse pose from the encoded fine-pose representation, supervised by the ground-truth coarse pose. The coarse loss $L_{coarse}$ can be defined as:
\begin{equation}
L_{coarse} = \sum_{t=1}^{T} \text{MSE}(CoarsePose_{pred}, CoarsePose_{gt}) .
\end{equation}

Then, we need to predict the final generated sequence from the encoded fine pose and coarse pose. We use the sequence $\hat{s}_t$ (referred to as $\text{FinePose}_{pred}$) obtained through fine-pose prediction in each refinement step, supervised by the ground truth Fine Pose to generate the final sequence. The loss can be defined as:
\begin{equation}
L_{refinement} = \ln(T)^2 \sum_{t=1}^{T} L_p(s_t, \hat{s}_t) ,
\end{equation}
where $L_p$ is a masked Mean Squared Error function \cite{arkushin2023ham2pose}:
\begin{equation}
L_p(s_t, \hat{s}_t) = \frac{1}{N} \sum\limits_{j=0}^{N} \frac{1}{|K|} \sum\limits_{i=0}^{|K|} c_i[j](s_t^i[j] - \hat{s}_t^i[j])^2 .
\end{equation}
Here, $N$ is the total number of frames in the sequence, $K$ is the set of keypoints, $c_i[j]$ is the confidence weight for the $i$-th keypoint in the $j$-th frame, and $s_t^i[j]$ and $\hat{s}_t^i[j]$ are the ground truth and predicted positions respectively. The ground truth $s_t$ at step $t$ is defined as an interpolation between the real sequence $s_0$ and the previous prediction $s_{t-1}$:
\begin{equation}
s_t = \delta_t s_0 + (1 - \delta_t) s_{t-1} ,
\end{equation}
where $\delta_t$ is the step size that increases with each refinement step.

The final composite loss function is a weighted sum of these components:
\begin{equation}
L_{total} = L_{refinement} + L_{len} + L_{coarse} .
\end{equation}

\section{Experiments and Results}
\label{sec:results}

\subsection{Implementation Details}



KANMultiSign is implemented on top of a Transformer encoder framework  \cite{vaswani2017attention}  with a hidden dimension of \(D = 128\) for encoding the HamNoSys text, step numbers, and positional information for both text and pose. A key feature of our architecture is a multi-scale, coarse-to-fine generation process, where the model first predicts a simplified 25-part coarse skeleton before regressing the final 137-keypoint fine pose. In addition, the FFN sublayer in the text-pose encoder is replaced by a KAN module \cite{liu2024kan}. This allows us to evaluate the effectiveness of KAN as a parameter-efficient alternative to standard MLPs for this task. All baseline models and ours were trained for up to 2,000 epochs using the Adam optimizer with a learning rate of \(1e-3\) and a batch size of 16 \cite{adam2014method}. The iterative refinement process consists of 10 steps. During training, we applied teacher forcing with a probability of 0.5 and injected Gaussian noise (\(\epsilon=1e-4\)) to stabilize learning. The total objective function was the sum of the fine-pose, coarse-pose, and sequence-length losses, with the sequence-length term weighted by \(\lambda = 2 \times 10^{-5}\). We implement KAN-FFN using FasterKAN \cite{Athanasios2024}. 
Unless otherwise stated, we keep the KAN basis configuration fixed across all experiments: the grid is defined on the interval $[-2, 2]$ with 8 grid points, and the basis-shaping hyperparameters are set to exponent 2 and denominator 0.33. 
We enable the base-update branch with a SiLU base activation, and initialize spline weights with a scale of 0.667. For reproducibility, all experiments were conducted on an NVIDIA GeForce RTX 4090D GPU with a fixed random seed of 42.

\begin{figure*}[htbp]
\centering
\includegraphics[width=\linewidth]{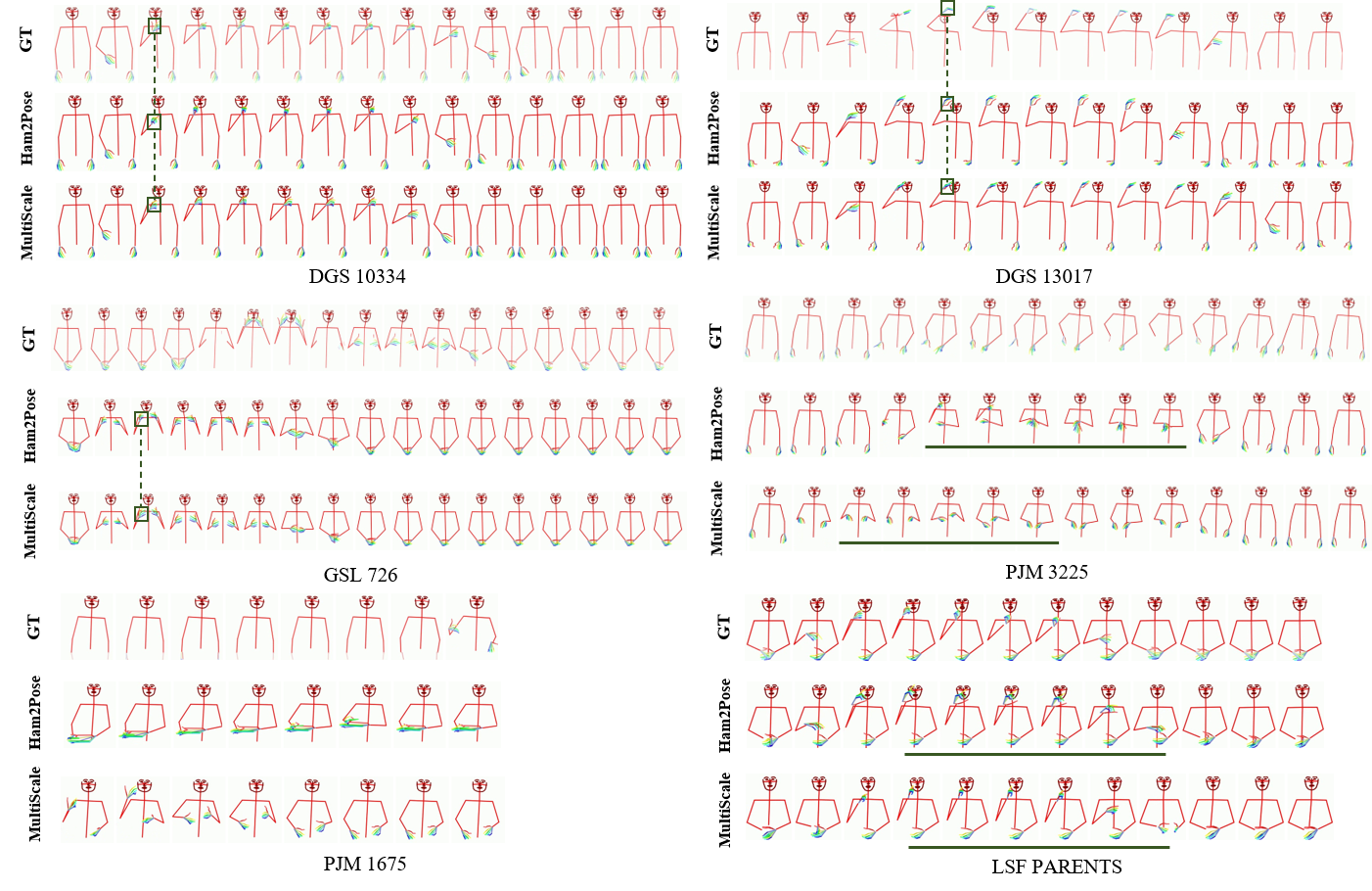}
\caption{Qualitative comparison on four signs from the DGS, GSL, LSF, and PJM datasets. Each example compares the ground truth, the baseline (Ham2Pose), and our proposed KANMultiSign. Our model demonstrates clear improvements in motion trajectory, handshape fidelity temporal dynamics, and spatial accuracy.} 
\label{fig:qualitative_comparison}
\end{figure*}

\subsection{Dataset}

We evaluate our proposed model on three datasets as mentioned earlier: DGS Corpus \cite{prillwitz2008dgs}, Dicta-Sign \cite{matthes2012dicta}, and the Corpus-based dictionary of Polish Sign Language (PJM) \cite{linde2014corpus}. These datasets collectively provide a diverse range of sign languages, including Polish Sign Language (PJM), German Sign Language (DGS), Greek Sign Language (GSL), and French Sign Language (LSF). Figure \ref{fig:sign_datasets} presents visual previews of sign sequences from all three datasets, showcasing representative examples from LSF, GSL, DGS, and PJM. To process these videos, we use the OpenPose \cite{cao2017realtime} pose estimation model, which annotates human poses with keypoints indicating their 2D positions and confidence levels. 

Each video frame includes 137 keypoints: 25 for the body ($K_B$), 70 for the face ($K_F$), and 42 for the hands ($K_H$, 21 per hand). To ensure data quality, we apply a confidence threshold, $c_{min}=0.2$, below which keypoints are considered unreliable. Frames are excluded if they fail the following criteria, ensuring only high-confidence data is used for model training:

\begin{equation} 
\begin{multlined} 
\quad \sum_{i\in{K_F}} c_i \leq c_{min}\cdot|K_F| \quad \textrm{or} \ c_{r_{wrist}}+c_{l_{wrist}} \leq c_{min}. 
\end{multlined} \label{eq} 
\end{equation}

This preprocessing step filters out frames with low facial keypoint confidence or poorly identified hand gestures, maintaining the integrity of the data for effective model evaluation.

\subsection{Qualitative Results}

\begin{figure}[htbp]
    \centering
    \includegraphics[width=0.95\linewidth]{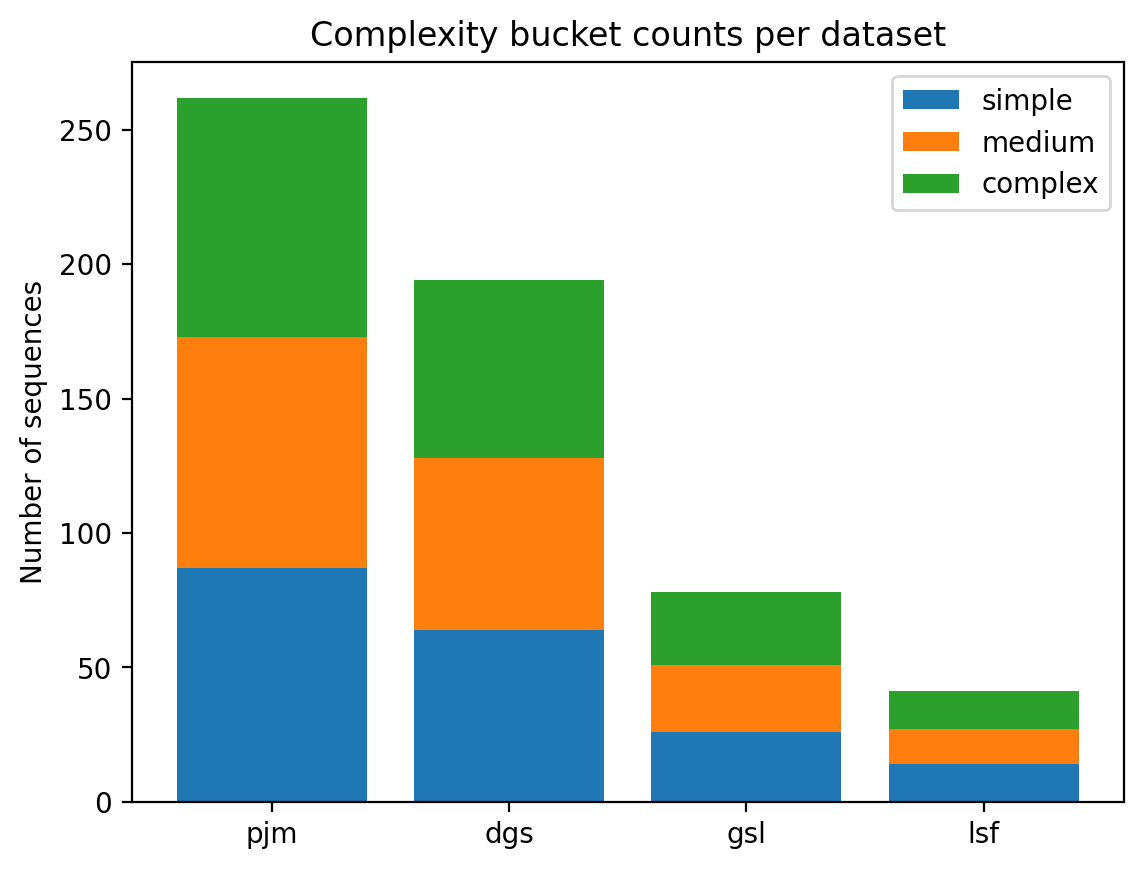}
    \caption{Complexity-bucket distribution on the test sets of PJM/DGS/GSL/LSF. Buckets (simple/medium/complex) are formed by per-dataset terciles of the complexity score. }
    \label{fig:bucket_distribution}
\end{figure}

To make the qualitative comparisons more systematic and to ensure coverage over different motion complexities, we stratify the test sequences into three complexity buckets (simple/medium/complex) within each dataset (Figure ~\ref{fig:bucket_distribution}). The complexity score combines (i) sequence length, (ii) mean per-frame keypoint displacement (motion magnitude), and (iii) mean second-order displacement (a jerk proxy capturing rapid motion changes). We then split sequences by the 33rd/66th percentiles of this score to form the three buckets and sample examples from each bucket. We report the bucket distribution per dataset to ensure coverage across gesture complexities. 

To stratify qualitative examples and reduce selection bias, we compute a per-sequence complexity score from 2D keypoints.
Given a sequence of length $T$ with keypoints $\mathbf{p}_{t,k}\in\mathbb{R}^2$ (frame $t$, keypoint $k$), we define
the motion magnitude
\begin{equation}
E = \frac{1}{T-1}\sum_{t=2}^{T}\frac{1}{|\Omega_t|}\sum_{k\in\Omega_t}\|\mathbf{p}_{t,k}-\mathbf{p}_{t-1,k}\|_2,
\end{equation}
and a jerk proxy capturing rapid motion changes
\begin{equation}
J = \frac{1}{T-2}\sum_{t=3}^{T}\frac{1}{|\Omega_t|}\sum_{k\in\Omega_t}\|(\mathbf{p}_{t,k}-\mathbf{p}_{t-1,k})-(\mathbf{p}_{t-1,k}-\mathbf{p}_{t-2,k})\|_2,
\end{equation}
where $\Omega_t$ denotes the set of valid keypoints at frame $t$ (i.e., keypoints with available confidence).
The final score is $C = z(\log(1+T)) + z(\log(1+E)) + z(\log(1+J))$, where $z(\cdot)$ denotes z-normalization.
Buckets are formed \emph{within each dataset} using terciles (33rd/66th percentiles) of $C$, yielding simple/medium/complex groups.

To visually demonstrate the outputs of our proposed model while avoiding selection bias, we report qualitative comparisons under the stratified complexity-bucket protocol described above. Specifically, for each dataset we partition the test set into three buckets and then select representative sequences from each bucket. Figures \ref{fig:qualitative_comparison} show example comparisons between the ground truth poses, the baseline Ham2Pose predictions, and KANMultiSign outputs. In each example, the ground truth is shown in the top row, the baseline prediction in the middle row, and the KANMultiSign output in the bottom row.

\textbf{DGS 10334 (Complex):} The ground truth sequence depicts a distinct brushing motion. However, the prediction of the baseline incorrectly orients the thumb towards the face and executes a simple shaking of the arm, failing to capture the intended trajectory. In contrast, our KANMultiSign model accurately predicts the correct hand orientation and successfully performs the brushing motion as described by the HamNoSys input. This demonstrates improved generation capability, likely originating in the global and joint-level context provided by the coarse supervision module.

\textbf{DGS 13017 (Medium):} In this ground truth sequence, the hand maintains a curled posture while moving toward the left side of the face. The baseline model not only fails to render the hand's curled shape accurately but also incorrectly positions the entire gesture to the right of the face. KANMultiSign exhibits an enhanced handshape that is more faithful to the ground truth. Although the full leftward movement is not perfectly replicated, our model correctly positions the hand centrally in front of the face, a significant spatial improvement over the baseline.

\textbf{GSL 726 (Complex):} The ground truth for this sign involves a rapid and symmetrical motion in which both hands move from the chin down to the chest, rotating the palms to face downward. The Ham2Pose baseline successfully captures the general downward trajectory but generates slightly curled fingers and fails to reproduce the quick temporal dynamic specified in the notation. However, KANMultiSign displays a more accurate hand configuration and successfully executes the required swift downward movement, better reflecting the intended dynamics of the sign.

\textbf{PJM 3225 (Complex):}This complex sign begins with both hands below the stomach performing symmetric pinching and touching motions. The baseline model generates a completely erroneous trajectory, first incorrectly raising the hands to the neck and crossing them at the chest before moving them outward. The prediction from KANMultiSign corrects these fundamental errors in both hand and arm positioning. Our model initiates the gesture in the correct spatial region below the stomach and performs the subsequent motions, aligning far more closely with the ground truth sequence.

\textbf{PJM 1675 (Simple):} The gesture starts from a neutral stance with both arms relaxed alongside the torso. The right forearm then lifts into the neutral signing space in front of the upper body and performs a brief, compact movement while forming a “5” handshape, before returning to a stable posture. Compared with this ground-truth pattern, the Ham2Pose baseline is disrupted by missing handshape cues and collapses into a confused low-position configuration, with limited range of hand motion and noticeable noise. Our MultiScale/KANMultiSign prediction produces a more plausible dynamic evolution by introducing the initial arm-lift and a more coherent hand action; however, it also generates residual extra hand motion in later frames beyond the ground truth.

\textbf{LSF PARENTS (Simple): } The gesture starts with the hand positioned near the chin, forming an indicative handshape with the index finger extended. It then makes a short lateral movement at chin level accompanied by a slight wrist rotation. Finally, the hand drops downward, during which the thumb and index finger gradually move closer and close together. In this example, the Ham2Pose baseline and our KANMultiSign output perform similarly overall: both produce reasonable hand-position changes in the correct spatial region and reproduce the main motion contour of the sign, without obvious trajectory drift or large pose errors.

These visual comparisons demonstrate that while our KANMultiSign model does not perfectly replicate the ground truth in all cases, it consistently shows improvements over the baseline Ham2Pose model. The enhancements are particularly evident in hand shape accuracy, gesture direction, and overall pose positioning. These results align with and visually support the quantitative improvements discussed in the previous sections.

\begin{table*}[htbp]
\centering
\caption{The comparison of DTW-MJE and nDTW-MJE metrics}
\begin{tabular}{c|c|c|c|c|c|c|c|c|c}
\toprule
\multirow{2}{*}{} & \multicolumn{4}{c|}{DTW-MJE} & \multicolumn{4}{c|}{nDTW-MJE} & \multirow{2}{*}{Param Size $\downarrow$} \\
\cline{2-9}
 & PJM $\downarrow$ & DGS $\downarrow$ & GSL $\downarrow$ & LSF $\downarrow$ & PJM $\downarrow$ & DGS $\downarrow$ & GSL $\downarrow$ & LSF $\downarrow$ & \\
\hline
Ham2Pose & 33.62 & 19.45 & 25.61 & 14.23 & \textbf{8.43} & 10.10 & 15.73 & 9.83 & 3.8M \\
\hline
\textbf{MultiSign (Ours)} & 29.62 & \textbf{18.50} & \textbf{24.02} & \textbf{13.68} & 8.98 & 10.30 & \textbf{15.26} & \textbf{9.60} & 3.8M\\
\hline
\textbf{\makecell{KANMultiSign(Ours)}} & \textbf{29.56} & 18.52 &25.14 & 14.57 & 8.51 & \textbf{9.95} & 15.58 & 10.26 & \textbf{1.7M} \\
\bottomrule
\end{tabular}
\label{table:comparison}
\end{table*}

\subsection{Quantitative Results}

Table \ref{table:comparison} presents a comparison of our MultiSign model against the baseline Ham2Pose model \cite{arkushin2023ham2pose} using DTW-MJE and nDTW-MJE metrics across different sign languages. The results show that the clearest and most consistent DTW-MJE improvements come from the introduction of the coarse-to-fine multi-scale supervision pathway. In particular, the standard MultiSign model improves DTW-MJE across all four sign languages relative to Ham2Pose, indicating that the multi-scale design is the main driver of motion-accuracy gains. The KAN-based variant achieves a substantial reduction in model size, decreasing parameters from 3.8M to 1.7M, while remaining broadly competitive under the same multi-scale framework. Its quantitative results are comparable to the standard multi-scale model on several datasets, although it does not uniformly deliver the strongest scores across all metrics and languages. We therefore interpret the role of KAN primarily as improving parameter efficiency, rather than as the dominant source of accuracy improvement.

\begin{table*}[htbp]
\centering
\caption{Evaluation metrics (Distance Rank). The table shows Top-K accuracy for generated-to-ground-truth retrieval (Pred) and ground-truth-to-generated retrieval (Label). Higher values are better.}
\begin{tabular}{lccc}
\toprule
\multirow{2}{*}{Model} & \multicolumn{3}{c}{Pred / Label} \\
\cmidrule(l){2-4} 
 & \textbf{Rank 1 $\uparrow$} & \textbf{Rank 5 $\uparrow$} & \textbf{Rank 10 $\uparrow$} \\
\midrule
Ham2Pose & 0.19 / 0.51 & \textbf{0.59} / 0.79 & 0.81 / 0.88 \\
\textbf{MultiSign (Ours)} & 0.20 / 0.50 & \textbf{0.59} / 0.79 & \textbf{0.85} / 0.86 \\
\textbf{KANMultiSign (Ours)} & \textbf{0.21} / \textbf{0.53} & 0.57 / \textbf{0.82} & \textbf{0.85} / \textbf{0.89}\\
\bottomrule
\end{tabular}
\label{table:evaluation_metrics}
\end{table*}

To further evaluate the quality and distinctiveness of our generated pose sequences, we employ the Distance Rank metric. Table \ref{table:evaluation_metrics} shows the results of this evaluation. In the prediction-to-ground-truth (pred) comparison, our MultiScale model outperforms the baseline in all three rank categories (1, 5, and 10), indicating that our generated sequences are more consistently similar to their corresponding ground truth. For the Ground Truth to Prediction (label) comparison, the performance is comparable to the baseline, with slightly lower Rank 5 and 10 scores but similar Rank 1 scores. And KANMultiSign also shows strong performance across most rank categories.

These results demonstrate that our multi-scale model enhances the spatial accuracy of generated pose sequences, as evidenced by the improved DTW-MJE scores. The model also maintains or slightly improves the distinctiveness of generated sequences within the broader context of the dataset, as shown by the Distance Rank metrics. The mixed results in nDTW-MJE suggest that while KANMultiSign generally improves spatial accuracy, there might be a trade-off in handling missing keypoints for some sign languages. This could be an area for future investigation and improvement.

Overall, these results indicate that our multi-scale approach offers meaningful improvements in sign language pose generation, particularly in terms of spatial accuracy and sequence distinctiveness, while maintaining comparable performance in handling missing keypoints.

\subsection{Ablation Study}

To systematically evaluate the individual contributions of our architectural choices, we conduct a comprehensive ablation study. 
We focus on three factors: (i) the coarse-to-fine multi-scale supervision pathway, (ii) replacing the standard MLP feed-forward networks (FFNs) in the text-pose encoder with KAN modules, and (iii) the depth of the KAN-based pose encoder. 
To ensure clear attribution, we first report a controlled 2$\times$2 ablation (Table~\ref{tab:ablation_2x2_results}) that disentangles FFN type and supervision scheme under the same backbone. 
We then provide extended ablations over additional configurations (Fig.~\ref{fig:ablation_study}), followed by a dedicated depth study for KAN encoders (Fig.~\ref{fig.supplementary_ablation_study}).

\begin{table*}[t]
\centering
\small
\setlength{\tabcolsep}{5.5pt}
\renewcommand{\arraystretch}{1.18}
\caption{2$\times$2 ablation on four datasets (PJM/GSL/DGS/LSF): FFN type (MLP vs.\ KAN) $\times$ training scheme (single-scale vs.\ coarse-to-fine multi-scale supervision). 
All settings share the same backbone except the ablated components: we use a 2-layer standard Transformer as the text encoder for all rows; the text-pose encoder is a 4-layer standard Transformer whose FFN is either MLP or replaced by KAN as indicated by each setting; and the pose projection is fixed to a linear layer  for all rows. Lower is better for both DTW and nDTW.}
\label{tab:ablation_2x2_results}
\begin{tabular}{l
                cccc
                cccc
                c}
\toprule
& \multicolumn{4}{c}{DTW} & \multicolumn{4}{c}{nDTW} & Params (M) \\
\cmidrule(lr){2-5}\cmidrule(lr){6-9}
Setting
& PJM & GSL & DGS & LSF
& PJM & GSL & DGS & LSF
& \\
\midrule
(A) MLP + single-scale
& 33.62 & 25.61 & 19.45 & 14.23
& 8.43  & 15.73 & 10.10 & 9.83
& 3.8 M\\
(B) MLP + coarse-to-fine
& 29.62 & 24.02 & 18.50 & 13.68
& 8.98  & 15.26 & 10.30 & 9.60
& 3.8 M\\
(C) KAN + single-scale
& 49.53 & 26.77 & 24.1 & 19.72
& 14.31  & 17.26 & 12.97  & 13.74
& 2.2 M\\
(D) KAN + coarse-to-fine
& 29.28 & 24.14 & 18.82 & 13.93
& 8.48  & 15.53 & 10.56 & 9.84
& 2.2 M \\
\bottomrule
\end{tabular}
\end{table*}

\subsubsection{The Effect of Core Architectural Components}
Table~\ref{tab:ablation_2x2_results} reports a controlled 2$\times$2 ablation that disentangles two key factors under the same backbone: (i) the supervision scheme (single-scale vs.\ coarse-to-fine multi-scale) and (ii) the FFN type in the text-pose encoder (MLP vs.\ KAN). 
Overall, coarse-to-fine supervision provides the most consistent gains: comparing (A) with (B), DTW improves across all four languages (e.g., PJM 33.62$\rightarrow$29.62; DGS 19.45$\rightarrow$18.50; LSF 14.23$\rightarrow$13.68), indicating that the multi-scale pathway is a primary driver of performance.
In contrast, replacing the FFN with KAN under single-scale supervision can degrade performance ((C) vs.\ (A), especially on PJM/LSF), suggesting higher sensitivity without intermediate constraints.
Crucially, combining KAN with coarse-to-fine supervision largely mitigates this degradation: (D) substantially improves over (C) (e.g., PJM DTW 49.53$\rightarrow$29.28; nDTW 14.31$\rightarrow$8.48) and achieves performance comparable to (B) with fewer parameters (2.2M vs.\ 3.8M), supporting KAN as a parameter-efficient alternative when coupled with multi-scale supervision.

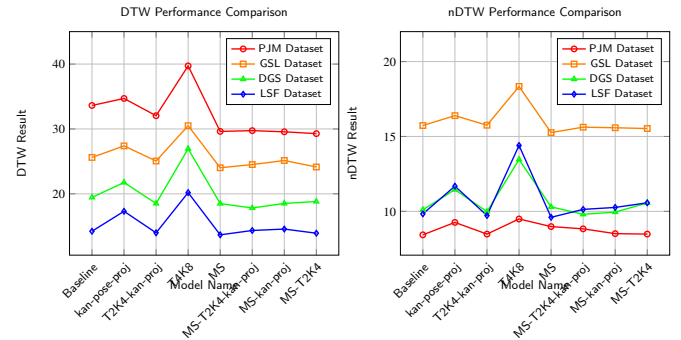
\begin{figure}[h]
\centering
\vspace{0.1cm}
\resizebox{0.5\textwidth}{!}{\begin{tikzpicture}
\begin{axis}[
   xlabel={Model Name},
   ylabel={DTW Result},
   symbolic x coords={Baseline, kan-pose-proj, T2K4-kan-proj, T4K8, MS, MS-T2K4-kan-proj, MS-kan-proj, MS-T2K4},
   xtick=data,
   xticklabel style={rotate=45, anchor=north east},
   legend pos=north east,
   grid=both,
   legend columns=1, 
   legend style={font=\footnotesize},
   title={DTW Performance Comparison},
   ymax=45
]
\addplot[color=red,mark=o,thick,line width=1pt] coordinates {
   (Baseline, 33.62) (kan-pose-proj, 34.70) (T2K4-kan-proj, 32.04) (T4K8, 39.72) (MS, 29.62) (MS-T2K4-kan-proj, 29.74) (MS-kan-proj, 29.56) (MS-T2K4, 29.28) 
};
\addlegendentry{PJM Dataset}
\addplot[color=orange,mark=square,thick,line width=1pt] coordinates {
   (Baseline, 25.61) (kan-pose-proj, 27.39) (T2K4-kan-proj, 25.06) (T4K8, 30.50) (MS, 24.02) (MS-T2K4-kan-proj, 24.51) (MS-kan-proj, 25.14) (MS-T2K4, 24.14) 
};
\addlegendentry{GSL Dataset}
\addplot[color=green,mark=triangle,thick,line width=1pt] coordinates {
   (Baseline, 19.45) (kan-pose-proj, 21.75) (T2K4-kan-proj, 18.51) (T4K8, 26.95) (MS, 18.50) (MS-T2K4-kan-proj, 17.82) (MS-kan-proj, 18.52) (MS-T2K4, 18.82) 
};
\addlegendentry{DGS Dataset}
\addplot[color=blue,mark=diamond,thick,line width=1pt] coordinates {
   (Baseline, 14.23) (kan-pose-proj, 17.31) (T2K4-kan-proj, 13.99) (T4K8, 20.19) (MS, 13.68) (MS-T2K4-kan-proj, 14.35) (MS-kan-proj, 14.57) (MS-T2K4, 13.93) 
};
\addlegendentry{LSF Dataset}
\end{axis}
\end{tikzpicture}
\vspace{0.1cm}
\begin{tikzpicture}
\begin{axis}[
   xlabel={Model Name},
   ylabel={nDTW Result},
   symbolic x coords={Baseline,  kan-pose-proj, T2K4-kan-proj, T4K8, MS, MS-T2K4-kan-proj, MS-kan-proj, MS-T2K4},
   xtick=data,
   xticklabel style={rotate=45, anchor=north east},
   legend pos=north east,
   grid=both,
   legend columns=1, 
   legend style={font=\footnotesize},
   title={nDTW Performance Comparison},
   ymax=22 
]
\addplot[color=red,mark=o,thick,line width=1pt] coordinates {
   (Baseline, 8.43) (kan-pose-proj, 9.26) (T2K4-kan-proj, 8.48) (T4K8, 9.49) (MS, 8.98) (MS-T2K4-kan-proj, 8.83) (MS-kan-proj, 8.51) (MS-T2K4, 8.48) 
};
\addlegendentry{PJM Dataset}
\addplot[color=orange,mark=square,thick,line width=1pt] coordinates {
   (Baseline, 15.73) (kan-pose-proj, 16.39) (T2K4-kan-proj, 15.75) (T4K8, 18.34) (MS, 15.26) (MS-T2K4-kan-proj, 15.62) (MS-kan-proj, 15.58) (MS-T2K4, 15.53) 
};
\addlegendentry{GSL Dataset}
\addplot[color=green,mark=triangle,thick,line width=1pt] coordinates {
   (Baseline, 10.1) (kan-pose-proj, 11.45) (T2K4-kan-proj, 9.98) (T4K8, 13.46) (MS, 10.3) (MS-T2K4-kan-proj, 9.8) (MS-kan-proj, 9.95) (MS-T2K4, 10.56) 
};
\addlegendentry{DGS Dataset}
\addplot[color=blue,mark=diamond,thick,line width=1pt] coordinates {
   (Baseline, 9.83) (kan-pose-proj, 11.68) (T2K4-kan-proj, 9.72) (T4K8, 14.4) (MS, 9.6) (MS-T2K4-kan-proj, 10.13) (MS-kan-proj, 10.26) (MS-T2K4, 10.569) 
};
\addlegendentry{LSF Dataset}
\end{axis}
\end{tikzpicture}}
\caption{Ablation study results comparing different model configurations across four sign language datasets. Performance is measured by DTW-MJE (left) and nDTW-MJE (right), where lower values indicate better performance. The model naming convention is detailed in the text.}
\label{fig:ablation_study}
\end{figure}

Beyond the controlled 2$\times$2 study in Table~\ref{tab:ablation_2x2_results}, we further report extended ablations in Fig.~\ref{fig:ablation_study} to evaluate additional configurations. For the extended ablations in Fig.~\ref{fig:ablation_study}, we adopt the following naming convention:

\begin{itemize}
    \item \textbf{MS}: Indicates the model uses the multi-scale Architecture.
    \item \textbf{TxKy}: Describes the encoder architecture, where $x$ is the number of layers in the text encoder and $y$ is the number of layers in the text-pose encoder. $T$ signifies a standard Transformer layer (with an MLP feed-forward network), and $K$ signifies our proposed encoder layer where the MLP is replaced by a KAN module. For example, \texttt{T2K4} denotes a model with a 2-layer Transformer text encoder and a 4-layer KAN-based text-pose encoder.
    \item \textbf{kan-proj}: Signifies that the final linear pose projection layer is replaced with a KAN.
\end{itemize}

\textbf{The Impact of Multi-Scale Architecture.} The results indicate that the most significant performance improvement comes from introducing the multi-scale supervision scheme. The \texttt{MS} model, which integrates this approach into the baseline's T2T4 architecture, consistently and substantially outperforms the \texttt{Baseline} across all four languages. For instance, in terms of DTW-MJE, the error on the PJM dataset drops from 33.62 to 29.62, and on the DGS dataset, it decreases from 19.45 to 18.50. A similar trend is observed for the LSF dataset (14.23 \(\rightarrow\) 13.68). This suggests that guiding the generation process with a coarse skeletal structure before refining fine-grained details is a highly effective strategy for improving overall spatial accuracy.

\begin{table*}[!htbp]
    \centering
    \caption{The effect of hidden dimension size in KAN layers. A smaller dimension of 64 shows better performance with fewer parameters.}
    \begin{tabular}{ccccc|c|cc|c}
        \toprule
         $E_{text}$ & $E_{text\ pose}$& MLP Size & Pose Proj. & MultiScale & $D_{hidden}$   & nDTW-MJE & DTW-MJE & Params \\
        \hline
         T T & K K K K & 2048  & K & X & 64  & 10.05 & 25.24 & 2.5M \\
         T T & K K K K & 2048  & K&  X& 128 & 11.37 & 35.43 & 3.0M \\
     \bottomrule
    \end{tabular}
    
    \label{tab:hidden_dim}
\end{table*}

\begin{table*}[!htbp]
    \centering
    \caption{The impact of text encoder and text-pose encoder configurations on model performance. T represents transformer layer, K represents KAN layer. Increasing model depth leads to worse performance and larger model size.}
    \begin{tabular}{cc|cccc|cc|c}
        \toprule
         $E_{text}$ & $E_{text\ pose}$ & $D_{hidden}$ & MLP Size & Pose Proj.& MultiScale & nDTW-MJE & DTW-MJE & Params \\
        \hline
         T T &  K K K K & 64  & 2048 & K & X& 10.05 & 25.24 & 2.5 M\\
         T T T T &  K K K K & 64  & 2048& K& X   & 11.06 & 26.51  & 3.7 M\\
         T T T T &  8K      & 64  & 2048& K& X   & 12.37 & 32.76 & 4.5 M\\
     \bottomrule
    \end{tabular}
\label{tab:encoder_config}
\end{table*}

\begin{table*}[!htbp]
    \centering
    \caption{The comparison of different text encoder configurations with fixed text-pose encoder. Using KAN layers (K) instead of transformer layers (T) in text encoder reduces model size while maintaining comparable performance.}
    \begin{tabular}{cc|cccc|cc|c}
        \toprule
         $E_{text}$ & $E_{text\ pose}$& MLP Size & $D_{hidden}$  & Pose Proj. & MultiScale  & nDTW-MJE & DTW-MJE & Params \\
        \hline
         K K & K K K K & 2048 & 64  & K& \checkmark & 10.08 & 24.16 & 1.7M \\
         T T & K K K K & - & 64 & K& \checkmark & 10.16 & 23.91 & 2.2M \\
     \bottomrule
    \end{tabular}
\end{table*}

\textbf{The Impact of KAN in Single-Scale Models.} When applied to the single-scale baseline architecture, the effectiveness of KAN appears to be nuanced. Using a deep KAN-based encoder (\texttt{T4K8}) or only replacing the projection layer (\texttt{kan-pose-proj}) leads to performance degradation compared to the \texttt{Baseline}. For example, the \texttt{T4K8} model increases the DTW error on PJM to 39.72. However, a more balanced architecture like \texttt{T2K4-kan-proj} outperforms the baseline on two datasets, reducing DTW error on DGS (18.51 vs. 19.45) and LSF (13.99 vs. 14.23), and improving nDTW on DGS (9.98 vs. 10.1) and LSF (9.72 vs. 9.83). This suggests that while KAN can be beneficial in a single-scale setting, their performance is highly sensitive to the architectural configuration.

\textbf{The Synergistic Effect of Multi-Scale and KAN.} 
The efficacy of the KAN-based modules is most evident when they are integrated into the multi-scale framework. All multi-scale variants incorporating KAN (\texttt{MS-T2K4-kan-proj}, \texttt{MS-kan-proj}, \texttt{MS-T2K4}) not only surpass the original \texttt{Baseline} by a large margin but also exhibit performance that is highly competitive with, and often superior to, the standard MLP-based \texttt{MS} model. Notably, specific KAN configurations achieve the best results on several datasets. The \texttt{MS-T2K4} model records the lowest DTW error on the PJM dataset (29.28). Furthermore, the \texttt{MS-T2K4-kan-proj} model excels on the DGS dataset, achieving the best overall DTW (17.82) and nDTW (9.8) scores among all tested configurations.

\subsubsection{The Effect of KAN Encoder Depth}
\begin{figure}[!htb]
\centering

\footnotesize
\resizebox{\linewidth}{!}{%
\begin{tabular}{lcccccccc}
\toprule
& \multicolumn{2}{c}{\textbf{PJM}} & \multicolumn{2}{c}{\textbf{GSL}} & \multicolumn{2}{c}{\textbf{DGS}} & \multicolumn{2}{c}{\textbf{LSF}} \\
\cmidrule(lr){2-3} \cmidrule(lr){4-5} \cmidrule(lr){6-7} \cmidrule(lr){8-9}
\textbf{Layers} & DTW & nDTW & DTW & nDTW & DTW & nDTW & DTW & nDTW \\
\midrule
\multicolumn{9}{l}{\textit{Without Multi-Scale Architecture}} \\
\quad 4 Layers & 49.53 & 14.31 & 26.77 & 17.26 & 24.10 & 12.97 & 19.72 & 13.74 \\
\quad 6 Layers & 34.81 & 9.33 & 27.69 & 16.86 & 21.33 & 11.16 & 15.64 & 10.61 \\
\quad 8 Layers & \textbf{32.82} & \textbf{8.54} & \textbf{25.06} & \textbf{15.69} & \textbf{19.03} & \textbf{10.12} & \textbf{14.62} & \textbf{10.43} \\
\midrule
\multicolumn{9}{l}{\textit{With Multi-Scale Architecture}} \\
\quad 4 Layers & \textbf{29.28} & \textbf{8.48} & \textbf{24.14} & \textbf{15.53} & 18.82 & 10.56 & 13.93 & 9.84 \\
\quad 6 Layers & 29.47 & 9.00 & 24.27 & 15.59 & \textbf{18.29} & \textbf{10.03} & \textbf{13.81} & \textbf{9.40} \\
\quad 8 Layers & 31.40 & 9.12 & 27.03 & 16.94 & 21.47 & 11.66 & 15.06 & 10.89 \\
\bottomrule
\end{tabular}
} %
\vspace{0.4cm} 

\begin{tikzpicture}
\begin{axis}[
    width=\linewidth,
    height=6cm,
    xlabel={Number of KAN Pose Encoder Layer},
    ylabel={DTW Result},
    symbolic x coords={4 Layer, 6 Layer, 8 Layer},
    xtick=data,
    grid=both,
    enlarge y limits={abs=2}
]
\addplot[color=red,mark=o,thick,line width=1pt] coordinates {
   (4 Layer, 49.53) (6 Layer, 34.81) (8 Layer, 32.82)
};
\addplot[color=red,mark=o,thick,line width=1pt,dashed] coordinates {
   (4 Layer, 29.28) (6 Layer, 29.47) (8 Layer, 31.40)
};
\addplot[color=orange,mark=square,thick,line width=1pt] coordinates {
   (4 Layer, 26.77) (6 Layer, 27.69) (8 Layer, 25.06)
};
\addplot[color=orange,mark=square,thick,line width=1pt,dashed] coordinates {
   (4 Layer, 24.14) (6 Layer, 24.27) (8 Layer, 27.03)
};
\addplot[color=green,mark=triangle,thick,line width=1pt] coordinates {
   (4 Layer, 24.10) (6 Layer, 21.33) (8 Layer, 19.03)
};
\addplot[color=green,mark=triangle,thick,line width=1pt,dashed] coordinates {
   (4 Layer, 18.82) (6 Layer, 18.29) (8 Layer, 21.47)
};
\addplot[color=blue,mark=diamond,thick,line width=1pt] coordinates {
   (4 Layer, 19.72) (6 Layer, 15.64) (8 Layer, 14.62)
};
\addplot[color=blue,mark=diamond,thick,line width=1pt,dashed] coordinates {
   (4 Layer, 13.93) (6 Layer, 13.81) (8 Layer, 15.06)
};
\end{axis}
\end{tikzpicture}

\vspace{0.2cm}
\begin{tikzpicture}
\begin{axis}[
    width=\linewidth,
    height=6cm,
   xlabel={Number of KAN Pose Encoder Layers},
   ylabel={nDTW Result},
   symbolic x coords={4 Layer, 6 Layer, 8 Layer},
    xtick=data,
    legend style={
        at={(0.5,-0.25)},  
        anchor=north,
        legend columns=2,
        font=\scriptsize,
        column sep=0.2cm,
        row sep=0.05cm,
        draw=black,
        fill=white,
        rounded corners=2pt,
        inner sep=0.1cm
    },
    grid=both,
    enlarge y limits={abs=2}
]
\addplot[color=red,mark=o,thick,line width=1pt] coordinates {
   (4 Layer, 14.31) (6 Layer, 9.33) (8 Layer, 8.54)
};
\addlegendentry{PJM (w/o Multi-Scale)}
\addplot[color=red,mark=o,thick,line width=1pt,dashed] coordinates {
   (4 Layer, 8.48) (6 Layer, 9.00) (8 Layer, 9.12)
};
\addlegendentry{PJM (w/ Multi-Scale)}
\addplot[color=orange,mark=square,thick,line width=1pt] coordinates {
   (4 Layer, 17.26) (6 Layer, 16.86) (8 Layer, 15.69)
};
\addlegendentry{GSL (w/o Multi-Scale)}
\addplot[color=orange,mark=square,thick,line width=1pt,dashed] coordinates {
   (4 Layer, 15.53) (6 Layer, 15.59) (8 Layer, 16.94)
};
\addlegendentry{GSL (w/ Multi-Scale)}
\addplot[color=green,mark=triangle,thick,line width=1pt] coordinates {
   (4 Layer, 12.97) (6 Layer, 11.16) (8 Layer, 10.12)
};
\addlegendentry{DGS (w/o Multi-Scale)}
\addplot[color=green,mark=triangle,thick,line width=1pt,dashed] coordinates {
   (4 Layer, 10.56) (6 Layer, 10.03) (8 Layer, 11.66)
};
\addlegendentry{DGS (w/ Multi-Scale)}
\addplot[color=blue,mark=diamond,thick,line width=1pt] coordinates {
   (4 Layer, 13.74) (6 Layer, 10.61) (8 Layer, 10.43)
};
\addlegendentry{LSF (w/o Multi-Scale)}
\addplot[color=blue,mark=diamond,thick,line width=1pt,dashed] coordinates {
   (4 Layer, 9.84) (6 Layer, 9.40) (8 Layer, 10.89)
};
\addlegendentry{LSF (w/ Multi-Scale)}
\end{axis}
\end{tikzpicture}
\caption{The comprehensive ablation study comparing the impact of KAN pose encoder layer depth with and without Multi-Scale structure. The architecture uses 2 Transformer text encoders with varying numbers of KAN pose encoder layers (4, 6, 8) and nn.Linear pose projection. Solid lines represent results without Multi-Scale structure, while dashed lines represent results with Multi-Scale structure. DTW performance is evaluated across four datasets: PJM, GSL, DGS, and LSF.}
\label{fig.supplementary_ablation_study}
\end{figure}
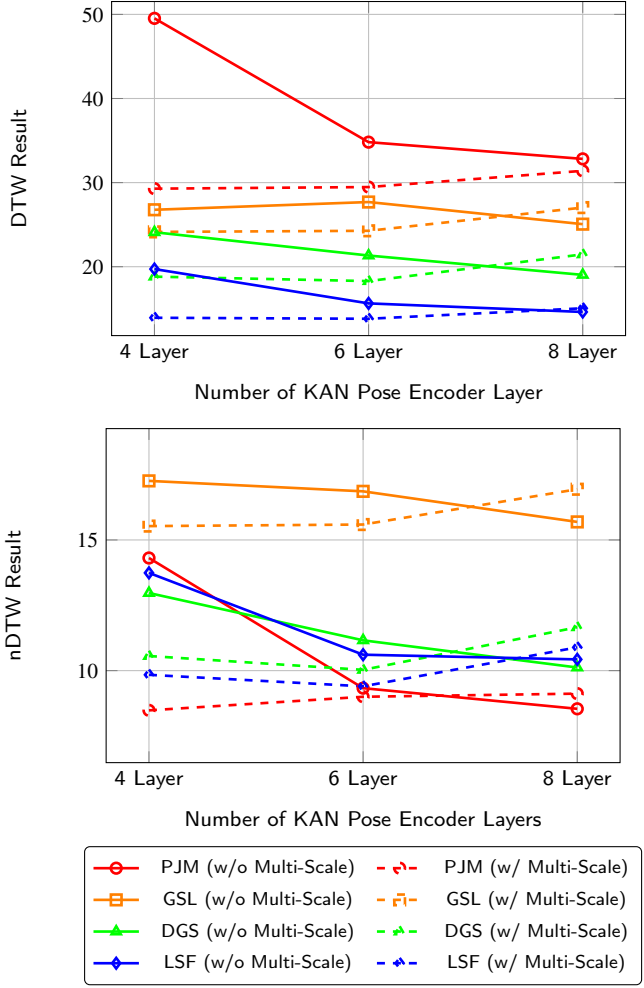

We further investigated the optimal configuration by analyzing the depth of the KAN pose encoder. We benchmarked models with 4, 6, and 8 KAN encoder layers, both with and without the multi-scale module. The results are presented in Figure~\ref{fig.supplementary_ablation_study}. models with varying numbers of KAN encoder layers (4, 6, and 8) with and without the multi-scale module, across the PJM, GSL, DGS, and LSF datasets. Performance was measured using DTW-MJE and nDTW-MJE, with the results presented in Figure \ref{fig.supplementary_ablation_study}.

The results clearly demonstrate that the multi-scale supervision architecture is the primary driver of performance. Its inclusion consistently leads to a significant reduction in both DTW and nDTW across all datasets, as indicated by the performance gap between the single-scale (solid lines) and multi-scale (dashed lines) models. For instance, on the PJM dataset, the 4-layer multi-scale model achieves a DTW of 29.28, decisively outperforming the best single-scale model (32.82 with 8 layers). This finding validates the effectiveness of our coarse-to-fine supervision strategy.

Furthermore, the study reveals a critical interaction between model depth and the multi-scale architecture. Without this architecture, performance consistently improves with a deeper encoder, with the 8-layer configuration yielding the best results across all datasets. In contrast, when the multi-scale framework is applied, shallower models become more effective. Specifically, the 4-layer encoder performed best on the PJM and GSL datasets, while the 6-layer encoder was optimal for DGS and LSF. Notably, the 8-layer configuration led to a decline in performance for all multi-scale models. This suggests that the multi-scale supervision simplifies the learning task, enabling a more compact model to achieve superior results. 

\begin{figure*}[!t]
\centering
\includegraphics[width=0.96\textwidth]{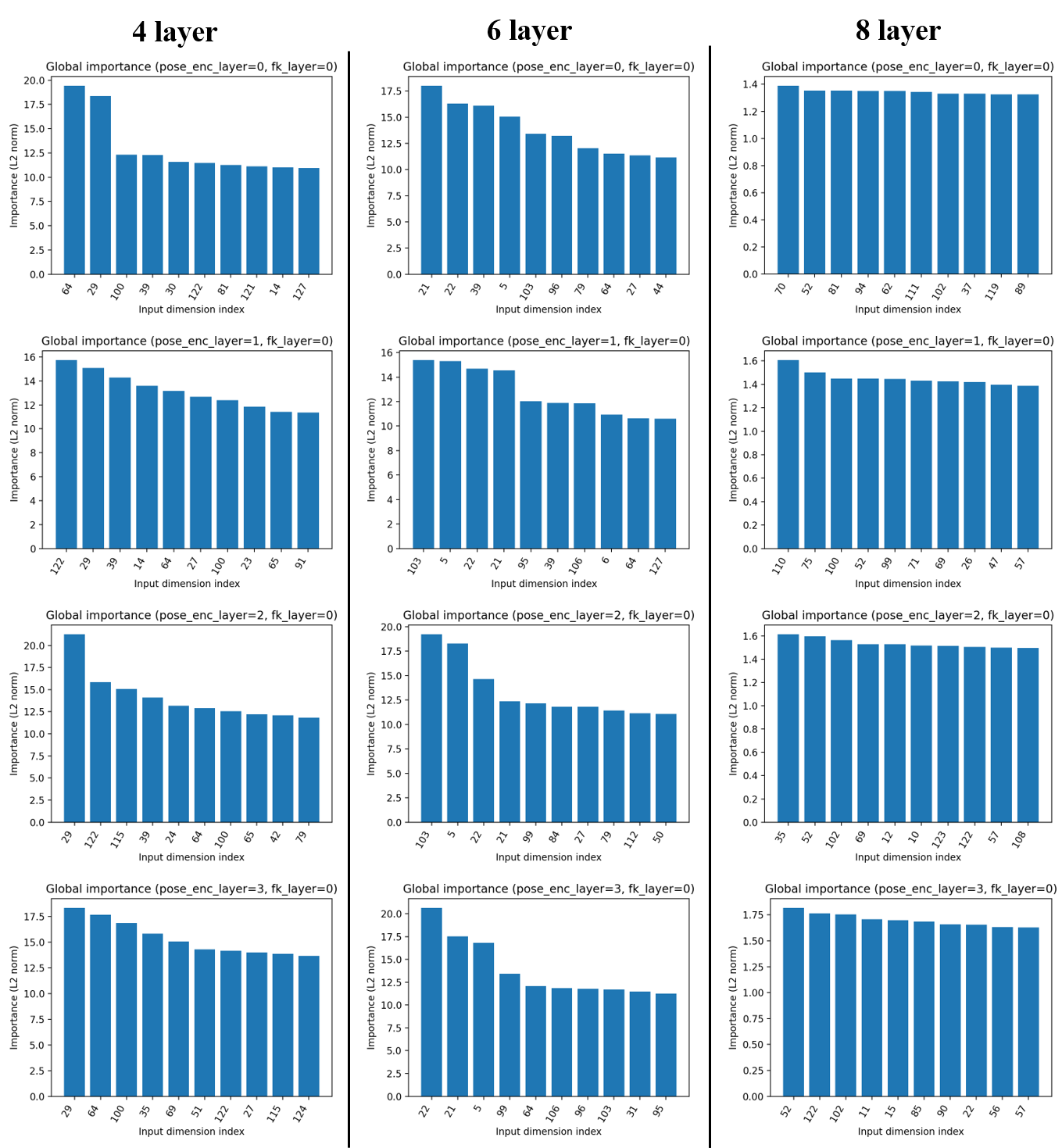}
\caption{\textbf{Global input-dimension importance of KAN-FFN vs.\ encoder depth. under MultiScale supervision.}
Columns correspond to pose-encoder depths of 4, 6, and 8 layers, respectively. Within each column, rows show the top-10 input dimensions ranked by global importance (measured by the L2 norm of spline parameters grouped by input channel) for pose-encoder layers 0 to 3. Shallow encoders (4/6 layers) exhibit more peaked and selective importance distributions, whereas the 8-layer encoder shows a flatter and lower-magnitude profile, suggesting reduced channel selectivity and increased redundancy when stacking many KAN-FFN blocks.
}

\label{fig:kan_depth_importance}
\end{figure*}

Figure~\ref{fig:kan_depth_importance} visualizes the global input-dimension importance of the KAN-FFN at each pose-encoder layer for different encoder depths (4/6/8 layers). We compute importance by grouping spline parameters by input channel and taking the L2 norm (top-10 dimensions shown per layer). A clear qualitative pattern emerges: for 4- and 6-layer encoders, the importance distributions are peaked, where a small subset of input dimensions consistently dominates across layers, indicating that the KAN-FFN learns more selective nonlinear primitives specialized to a few salient channels. In contrast, for the 8-layer encoder, the importance profile becomes noticeably flatter and less concentrated on a small set of dimensions, suggesting reduced channel selectivity and increased representational redundancy when stacking many KAN-FFN blocks. This observation is consistent with the non-monotonic depth trend under multi-scale supervision and supports our choice of a compact KAN encoder (4--6 layers) as the best accuracy--stability trade-off in the multi-scale setting.

Overall, the ablation results disentangle the sources of improvement and indicate that the primary accuracy gain is driven by the coarse-to-fine multi-scale supervision, which consistently reduces DTW-MJE when added to the baseline across the evaluated corpora. This is consistent with the interpretation that the intermediate coarse-pose pathway acts as a structural prior: it encourages the model to capture global kinematic consistency (e.g., torso and arm trajectories) before allocating capacity to fine-grained articulation. At the same time, KAN mainly contributes parameter efficiency, and its impact on accuracy is strongly conditioned on the presence of multi-scale constraints: replacing FFNs with KAN under single-scale supervision can be unstable and may degrade performance, whereas combining KAN with coarse-to-fine supervision largely mitigates this issue and remains competitive while using fewer parameters. A plausible explanation is that HamNoSys-to-pose generation involves highly nonlinear and heterogeneous mappings across joints and time; without coarse supervision to anchor global motion, the model can overfit local patterns or propagate early errors, which is particularly harmful under DTW-style evaluation. Therefore, the observed “better or comparable error with fewer parameters” is best explained by a synergistic effect: multi-scale supervision stabilizes the difficult symbol-to-pose mapping, while KAN provides a more compact nonlinear parameterization once sufficient intermediate constraints are available.

\subsection{Sensitivity Analysis}
We conducted a comprehensive sensitivity analysis to examine how different hyperparameters and architectural choices affect the model's performance. For this analysis, we focused on the overall nDTW-MJE and DTW-MJE metrics as key performance indicators.

\noindent\paragraph{The Impact of KAN Hidden Dimension.}
First, we investigated the impact of hidden dimensions in KAN layers. As shown in Table \ref{tab:hidden_dim}, we compared two configurations with hidden dimensions of 64 and 128, while keeping other parameters constant. The model with a hidden dimension of 64 achieved better performance (nDTW-MJE: 10.05, DTW-MJE: 25.24) compared to the 128-dimension version (nDTW-MJE: 11.37, DTW-MJE: 35.43). Additionally, the smaller dimension resulted in a more compact model with 2.5M parameters versus 3.0M parameters. This suggests that a smaller hidden dimension is sufficient to capture the necessary pose information, and increasing the dimension may lead to overfitting or unnecessary model complexity.

\noindent\paragraph{The Impact of Encoder Depth.}
Next, we analyzed various configurations of the text-pose encoder. The results demonstrate that the compact configuration achieves the best performance (nDTW-MJE: 10.05, DTW-MJE: 25.24) with the smallest parameter count (2.5M). Increasing the number of transformer layers in the text encoder to 4 led to worse performance (nDTW-MJE: 11.06, DTW-MJE: 26.51) and a larger model size (3.7M parameters). Further increasing the number of KAN layers to 8 resulted in even poorer performance (nDTW-MJE: 12.37, DTW-MJE: 32.76) and the largest model size (4.5M parameters). This suggests that a deeper architecture does not necessarily lead to better performance in this task, and the compact configuration provides an optimal balance between model capacity and performance.

\noindent\paragraph{The Impact of Text Encoder Architecture.}
Finally, we investigated the combined effects of Hamnosys encoder architecture types and MultiScale approach while maintaining consistent hidden dimensions and pose projection. The results show that replacing transformer layers with KAN layers in the HamNoSys text encoder and incorporating the MultiScale approach maintains comparable performance while significantly reducing model size. Specifically, the KAN-based configuration shows slightly better nDTW-MJE (10.08 compared to 10.16) while having marginally worse DTW-MJE performance (24.16 versus 23.91). Most importantly, the KAN-based configuration reduces the parameter count significantly from 2.2M to 1.7M, representing a 23\% reduction in model size.

The sensitivity analysis indicates that KANMultiSign’s performance depends more on appropriate capacity selection than on simply increasing depth. With other settings fixed, a smaller KAN hidden dimension (64) yields both lower errors and fewer parameters than a larger dimension (128), suggesting that increasing capacity can introduce unnecessary complexity and potential overfitting for this task. Similarly, expanding encoder depth does not monotonically improve results: the compact configuration provides the best accuracy–size trade-off, while deeper variants increase parameters and can worsen DTW/nDTW. Finally, replacing Transformer layers with KAN layers in the text encoder can preserve comparable accuracy while further reducing model size, reinforcing that KAN’s main advantage is parameter-efficient representation rather than brute-force scaling.

\subsection{KAN interpretability}
\begin{figure*}[t]
\centering
\scriptsize
\setlength{\tabcolsep}{-1.4pt}
\renewcommand{\arraystretch}{0.7}

\newcommand{\kanw}{0.249\textwidth}
\hspace*{-10pt}
\begin{tabular}{c c c c}
\includegraphics[width=\kanw]{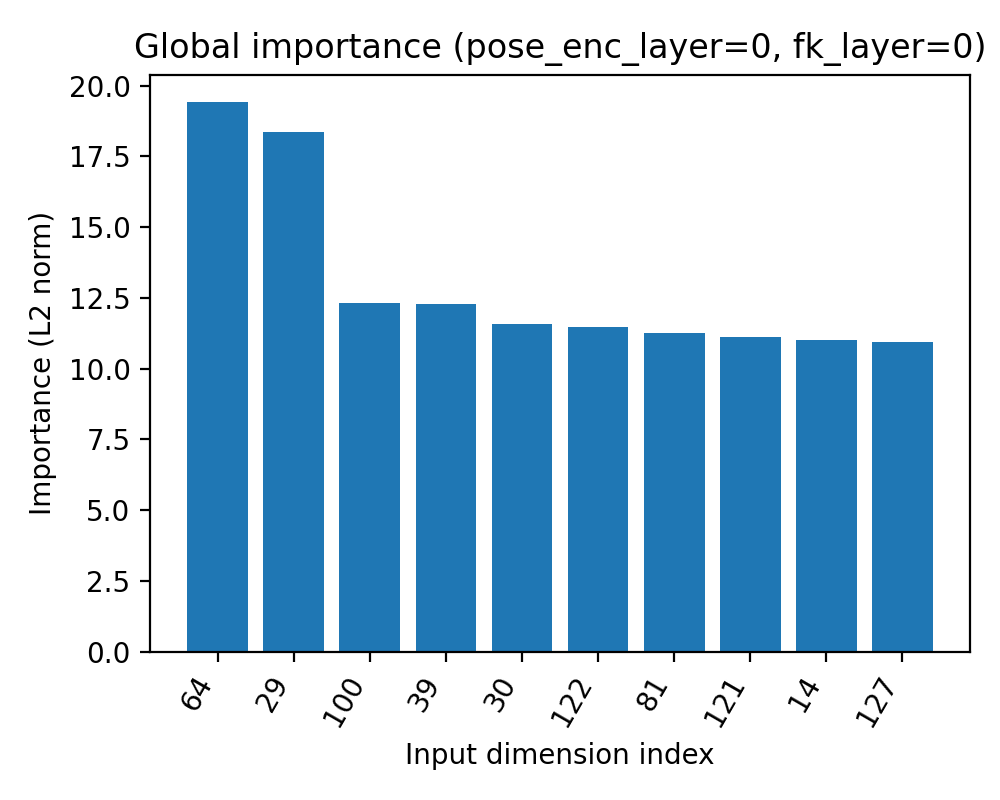} &
\includegraphics[width=\kanw]{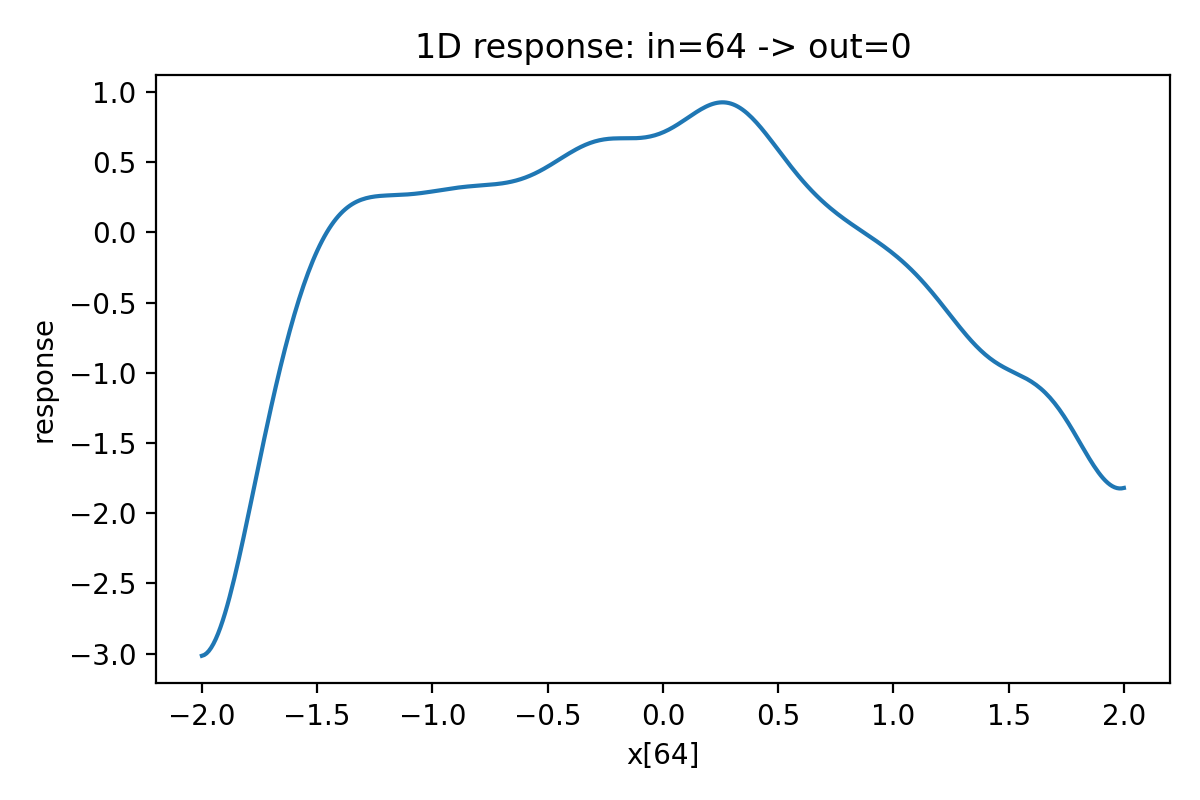} &
\includegraphics[width=\kanw]{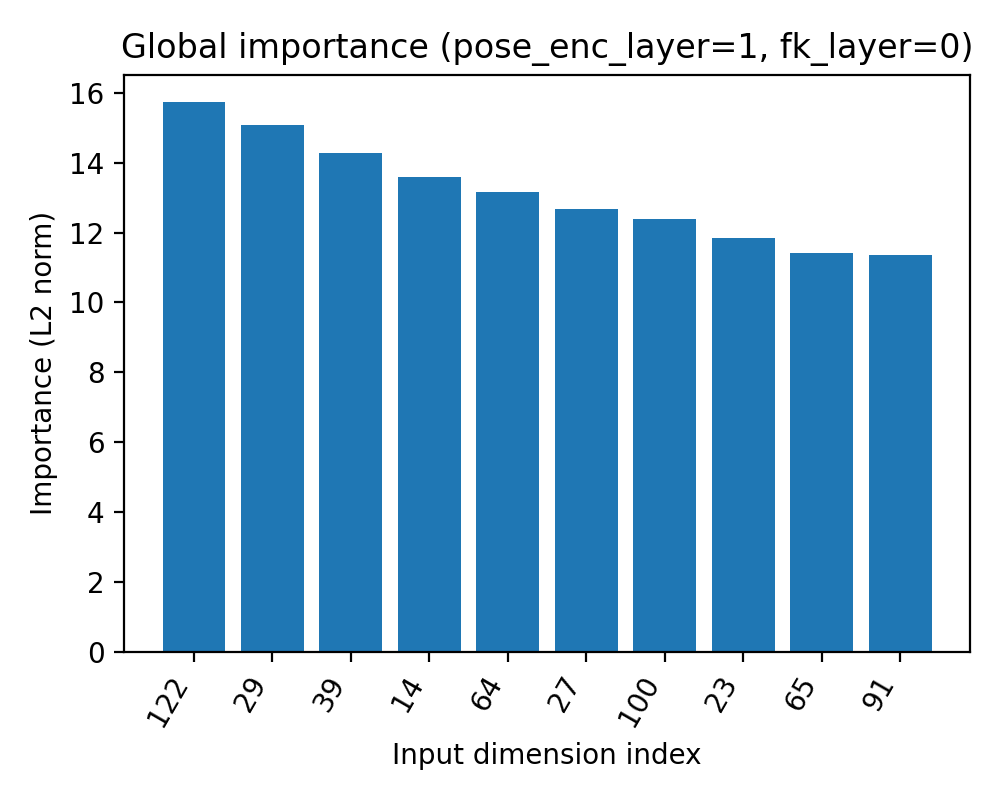} &
\includegraphics[width=\kanw]{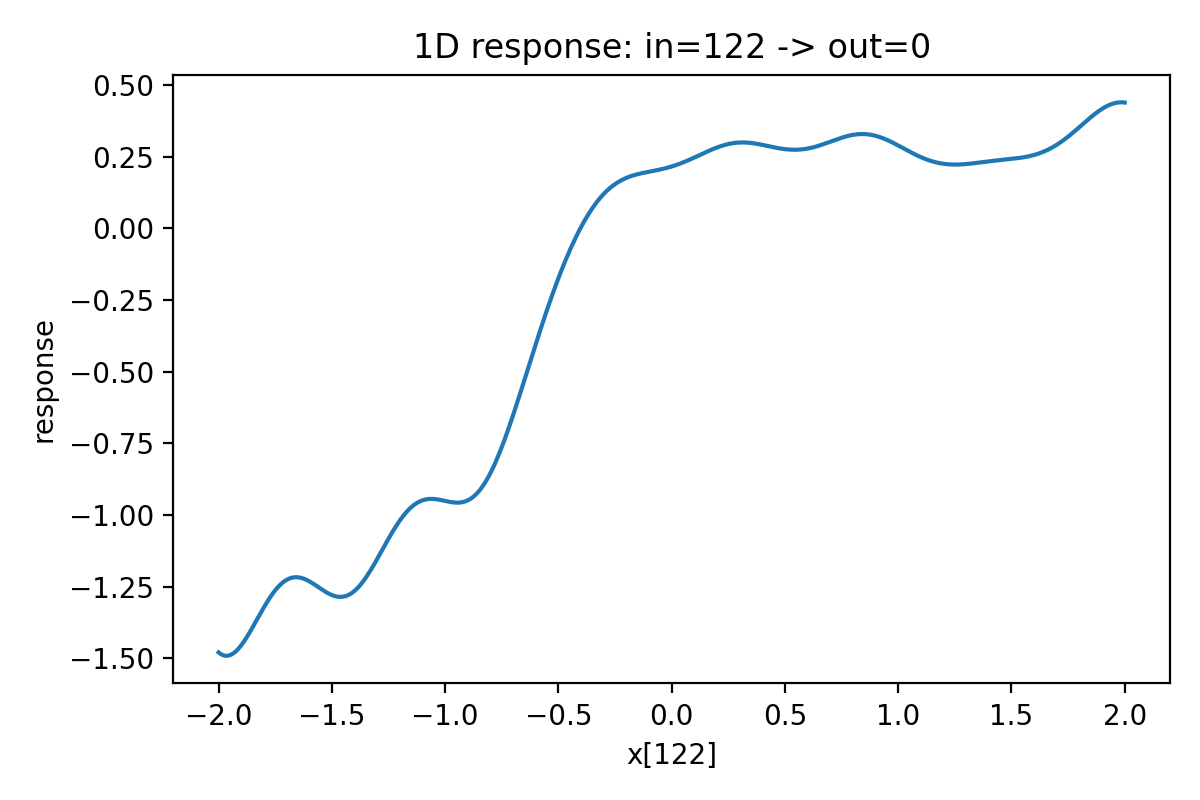} \\[-2mm]
\includegraphics[width=\kanw]{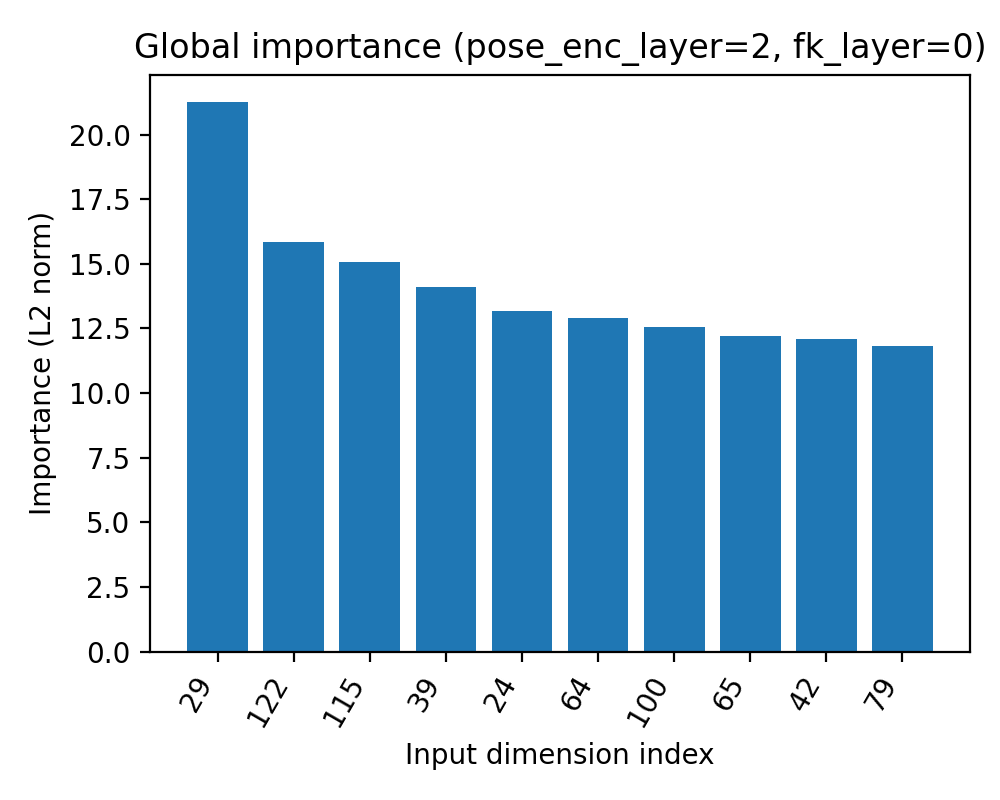} &
\includegraphics[width=\kanw]{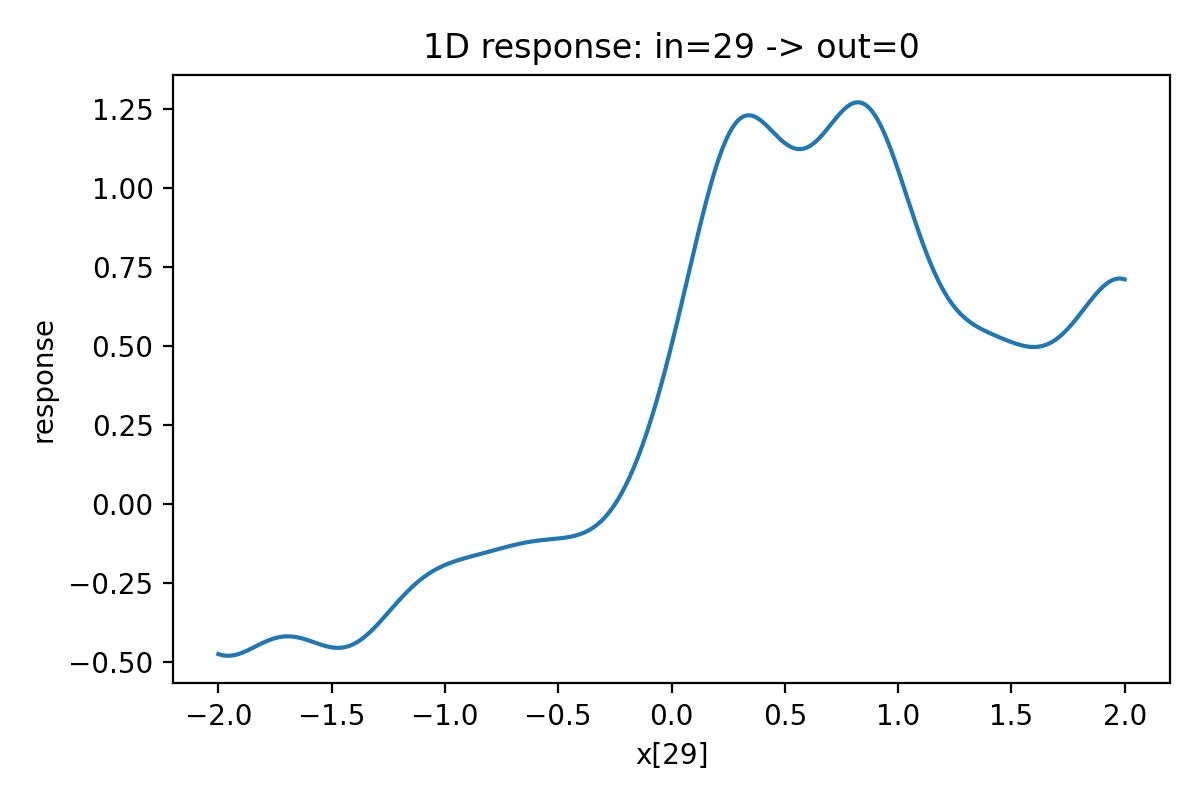} &
\includegraphics[width=\kanw]{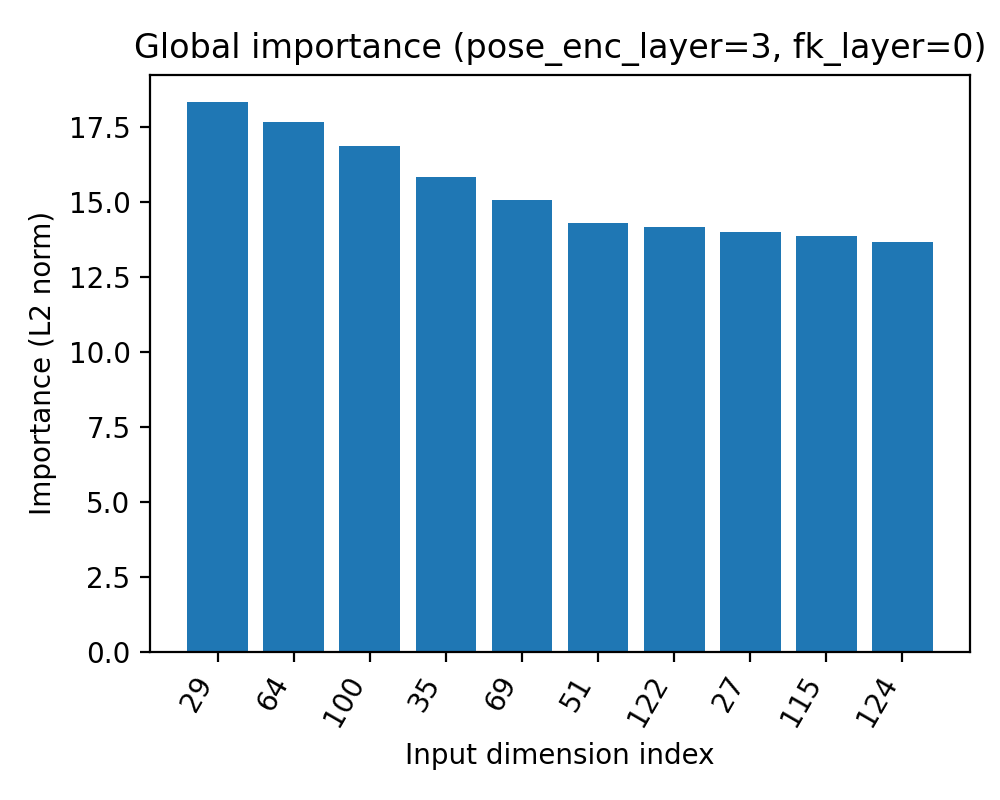} &
\includegraphics[width=\kanw]{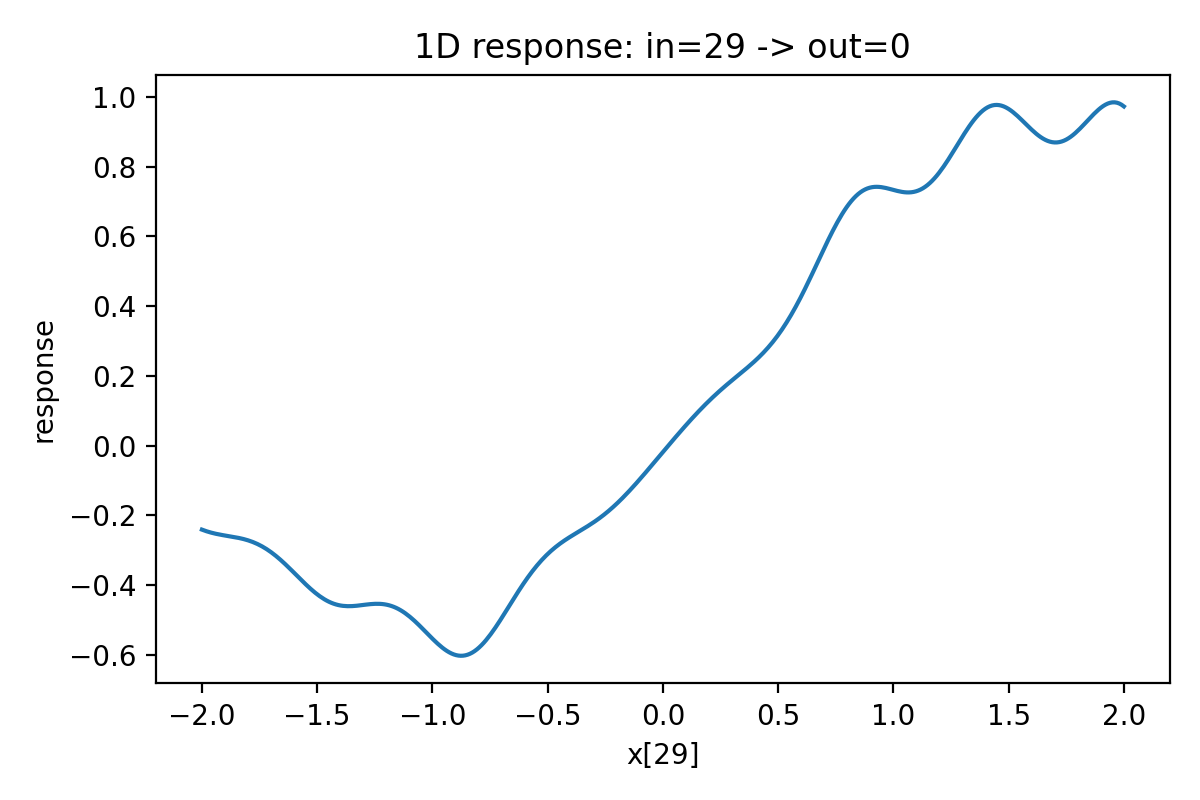} \\
\end{tabular}

\vspace{-2mm}
\caption{\textbf{Visualization of KAN-based FFNs in the trained multi-scale model.}
For each pose-encoder layer (00--03), we show grouped input-dimension importance and the learned one-dimensional response of the top-ranked input dimension in KANLayer-0. Importance is measured by the L2 norm of spline parameters aggregated by input channel. The one-dimensional curves visualize the intrinsic univariate mappings parameterized by KAN. LayerNorm is omitted here to avoid cross-dimension coupling in the interpretation.}
\label{fig:kan_interpretability}
\vspace{-3mm}
\end{figure*}

To address the interpretability of KAN, we visualize the learned univariate function primitives from the KAN-based FFN modules in the text-pose encoder (Figure ~\ref{fig:kan_interpretability}). We use a 2-layer standard Transformer as the text encoder, a 4-layer standard Transformer as the text-pose encoder whose FFN is replaced by KAN, and a fixed linear pose projection.
We compute input-dimension importance by grouping spline weights by input channel and taking the L2 norm over output channels and grid coefficients.
For each text-pose encoder layer, we plot a representative 1D response curve by varying only the top-ranked input dimension (rank-00) while fixing other dimensions to zero, which isolates the intrinsic univariate mapping parameterized by KAN (basis + spline weights).
LayerNorm is omitted in this visualization to avoid cross-dimension coupling and to keep the mapping strictly 1D.

\subsection{Sequence length}

\begin{figure}[t]
\centering
\includegraphics[width=0.48\linewidth]{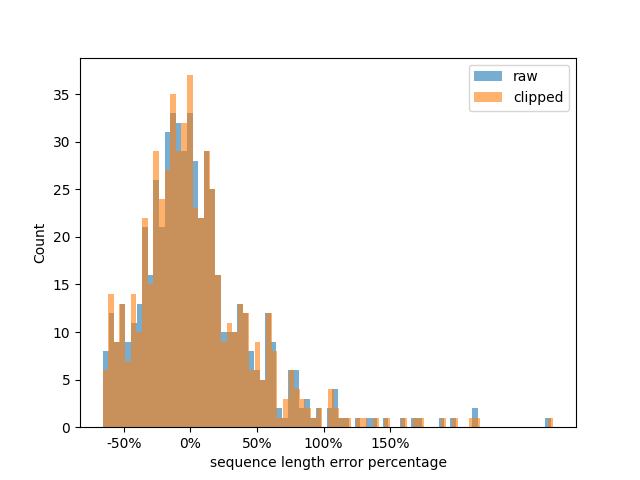}
\includegraphics[width=0.48\linewidth]{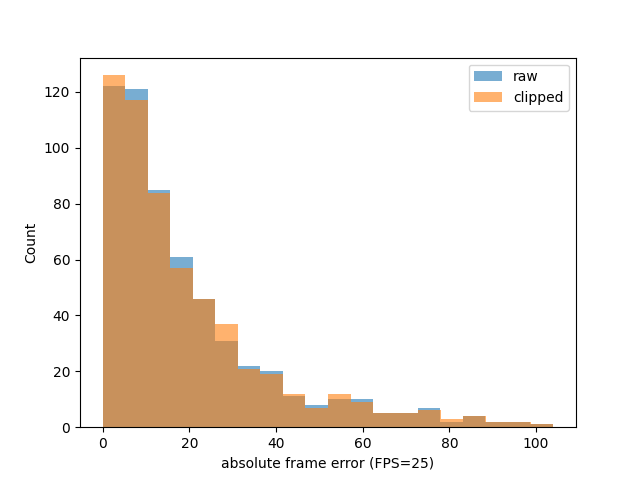}
\caption{Distributions of sequence-length prediction errors on the test split. We compare the raw output of the length head with the inference-time length used for generation (rounded and clipped to 20--200 frames). The two distributions closely overlap, indicating that clipping mainly enforces valid lengths rather than changing accuracy. Left: relative error in percent. Right: absolute error in frames (FPS=25).}
\label{fig:seq-len-hist}
\end{figure}

We evaluate the sequence-length predictor on the test split using MAE and MSE in frames (FPS=25) (Figure \ref{fig:seq-len-hist}). We report both the raw regressor output and the inference-time rounded+clipped length $\hat{l}_{\text{final}}$, which is the value used during generation (Tab.~\ref{tab:seq-len-error}). Overall, the clipped predictor achieves $\mathrm{MAE}=19.25$ frames and $\mathrm{MSE}=740.55$ frames$^{2}$ over $N=575$ samples. Raw and clipped statistics are nearly identical, indicating that clipping acts as a validity constraint rather than materially changing accuracy. Length under-estimation compresses motion dynamics, while over-estimation may introduce stretched trajectories or idle segments; both can degrade temporal 
alignment and increase DTW-based distances.

This length analysis also helps interpret cross-dataset variations in nDTW-MJE. 
Although DTW-MJE improves consistently, nDTW-MJE can still fluctuate across corpora (Table.~\ref{table:comparison}). 
One plausible factor is temporal alignment: length mismatch may induce global time scaling (motion compression or stretching), 
which can alter the DTW warping path and affect nDTW-based distances. 
As shown in Tab.~\ref{tab:seq-len-error}, GSL has the lowest length prediction error (MAE=17.80 frames), 
whereas LSF has the highest error (MAE=20.87 frames) and the smallest test set ($N=41$). 
Accordingly, nDTW-MJE improvements tend to be more stable on GSL, while on LSF the larger length error and small-$N$ regime 
may contribute to higher variance and occasional slight nDTW-MJE regressions, even when the overall spatial trajectory quality remains comparable.

\begin{table}[t]
\centering
\small
\setlength{\tabcolsep}{4pt}
\renewcommand{\arraystretch}{1.05}
\begin{tabular}{l r r r r r}
\hline
Dataset & $N$ & R-MAE$\downarrow$ & R-MSE$\downarrow$ & C-MAE$\downarrow$ & C-MSE$\downarrow$ \\
\hline
Overall & 575 & 19.23 & 739.52 & 19.25 & 740.55 \\
DGS & 194 & 18.39 & 805.59 & 18.40 & 807.10 \\
GSL & 78  & 17.80 & 594.60 & 17.79 & 594.56 \\
LSF & 41  & 20.87 & 645.85 & 20.88 & 644.49 \\
PJM & 262 & 20.02 & 748.39 & 20.05 & 749.76 \\
\hline
\end{tabular}
\caption{Sequence length prediction error on the test split (FPS=25). ``R'' denotes the raw continuous output of the sequence-length head. ``C'' denotes the inference-time length used for generation, obtained by rounding the raw prediction to an integer number of frames and clipping it to the valid range of 20--200 frames. MAE and MSE are computed in frames (and frames squared, respectively); lower is better.}
\label{tab:seq-len-error}
\end{table}

\subsection{Efficiency Analysis}

Table~\ref{tab:efficiency_comparison} reports efficiency metrics beyond parameter count, including training time per epoch, peak training GPU memory, inference latency, and peak inference GPU memory, all measured under the same hardware setting. The results show that efficiency trade-offs differ across model variants. The KAN-based MS-T2K4 model achieves the smallest parameter count (2.2M) and the lowest inference memory footprint (0.093 GB), confirming its advantage in compactness and memory-efficient inference. However, the standard multi-scale MLP-based model (MS-T2T4) achieves the lowest inference latency (19.81 ms) and the lowest training memory usage (3.139 GB), indicating that reduced parameter count does not necessarily translate into uniformly lower computational cost. Deeper KAN-based variants (MS-T2K6 and MS-T2K8) further increase both training cost and memory usage. Overall, these results suggest that the main efficiency advantage of the KAN-based variant lies in parameter reduction and lower inference memory, rather than consistent speedup across all efficiency metrics.

\section{Limitations and Future Work}
\begin{table*}[t]
\centering
\small
\caption{Efficiency comparison under the same hardware and batch setting (single NVIDIA RTX 4090D GPU, batch size = 16). Training cost is summarized by the average training time per epoch and the average peak training GPU memory across 20 epochs. Inference metrics are reported as the mean latency over repeated runs together with peak inference GPU memory. Lower is better for all metrics.}
\label{tab:efficiency_comparison}
\begin{tabular}{lccccc}
\toprule
Model & Time/Epoch (s) & Train Mem. (GB) & Latency (ms)& Infer. Mem. (GB)  & Params\\
\midrule
Ham2Pose & 4.43 & 3.249 & 35.88 & 0.122 & 3.8 M\\
MS-T2T4 & 4.49 & \textbf{3.139} & \textbf{19.81} & 0.100 & 3.8 M\\
MS-T2K4& \textbf{4.19} & 3.894 & 28.87 & \textbf{0.093} & \textbf{2.2M}\\
MS-T2K6& 5.30 & 4.675 & 42.00 & 0.103 & 2.6 M \\
MS-T2K8 & 6.24 & 6.132 & 37.74 & 0.105 & 3.0 M\\
\bottomrule
\end{tabular}
\end{table*}

A key limitation of our current model lies in its handling of missing keypoints during ground-truth coarse pose extraction. Our coarse targets are constructed by pooling fine-grained joints within each body part. While mean-pooling is generally stable, it may become biased when a subset of joints is missing; conversely, a naive confidence-weighted average may be numerically fragile under severe confidence sparsity.

A worthwhile future direction is to explore more robust coarse target construction when confidence is sparse. One possible approach is temporally consistent interpolation or smoothing of missing joints prior to pooling, which may reduce discontinuities in the coarse targets. In addition, improving the upstream keypoint extraction quality may also help: adopting a stronger pose estimator could reduce missingness and increase the reliability of both coarse targets and downstream supervision.

In addition, an important open question is generalization under limited training data. Evaluating and improving low-resource robustness (e.g., via controlled subsampling studies) is a valuable future direction, as notation-driven SLP is often most needed precisely when paired training material is scarce.

A further worthwhile direction is to compare KAN-based models with explicitly capacity-matched compressed MLP/Transformer baselines under similar parameter budgets. This would enable a more controlled assessment of whether the observed efficiency is attributable to the KAN formulation itself or to differences in model-capacity allocation.

Furthermore, the advent of Large Language Models (LLMs) presents a promising avenue. They could potentially offer a deeper semantic understanding of the symbolic notation, providing contextual cues that help resolve ambiguity, leading to the generation of more naturalistic and context-aware sign language animations.

\section{Conclusion}
\label{sec:con}

In this paper, we present KANMultiSign, a framework for generating sign language pose animations from symbolic HamNoSys notation. Our approach centers on a multi-scale, coarse-to-fine generation process that improves global structural coherence and fine-grained articulation, while also investigating parameter-efficient KAN modules as replacements for Transformer feed-forward sublayers. Experiments across four sign language corpora show that the multi-scale design is the primary source of motion-accuracy improvement over a strong baseline. Within this framework, KAN-based variants substantially reduce parameter count while maintaining competitive generation quality. We also discuss the limitations of our approach and outline directions for future work.

\bibliographystyle{elsarticle-num}
\bibliography{refs}

\end{document}